%% file: colm2024_conference.tex
\documentclass{article} %
\usepackage{colm2024_conference}

\usepackage{booktabs}
\usepackage{graphicx}
\usepackage{enumitem}
\usepackage{wrapfig}
\usepackage{algorithm}
\usepackage{algpseudocode}
\usepackage{wrapfig}
\usepackage{float}
\usepackage{microtype}
\usepackage{amsmath}
\usepackage{amssymb}
\usepackage{colortbl}
\usepackage[utf8]{inputenc}
\definecolor{lightgray}{rgb}{0.9,0.9,0.9}
\usepackage{caption}
\usepackage{subcaption}
\usepackage{xcolor}
\usepackage{setspace}
\usepackage{url}
\usepackage{multirow}
\usepackage{colortbl}
\usepackage{tabularx}
\usepackage{blindtext}
\usepackage{pgfplots}
\pgfplotsset{compat=1.18} 
\usepackage{tikz}
\usetikzlibrary{er,positioning,bayesnet}
\usepackage{makecell}
\usepackage{tipa}
\usepackage{siunitx}
\usepackage{nicefrac}
\usepackage{tocloft}
\usepackage{listings}
\usepackage[raster,skins]{tcolorbox} %
\usepackage{xltabular}
\usepackage{adjustbox}
\usepackage{xurl}
\input{math_commands.tex}

\makeatletter
\DeclareRobustCommand\onedot{\futurelet\@let@token\@onedot}
\def\@onedot{\ifx\@let@token.\else.\null\fi\xspace}

\makeatother

\title{Hunyuan3D Studio: End-to-End AI Pipeline for \\ Game-Ready 3D Asset Generation}

\author{
\bf Tencent Hunyuan3D 
}

\begin{document}

\maketitle

\begin{figure}[ht]
    \vspace{-10pt}
    \centering
    \includegraphics[width=1\linewidth]{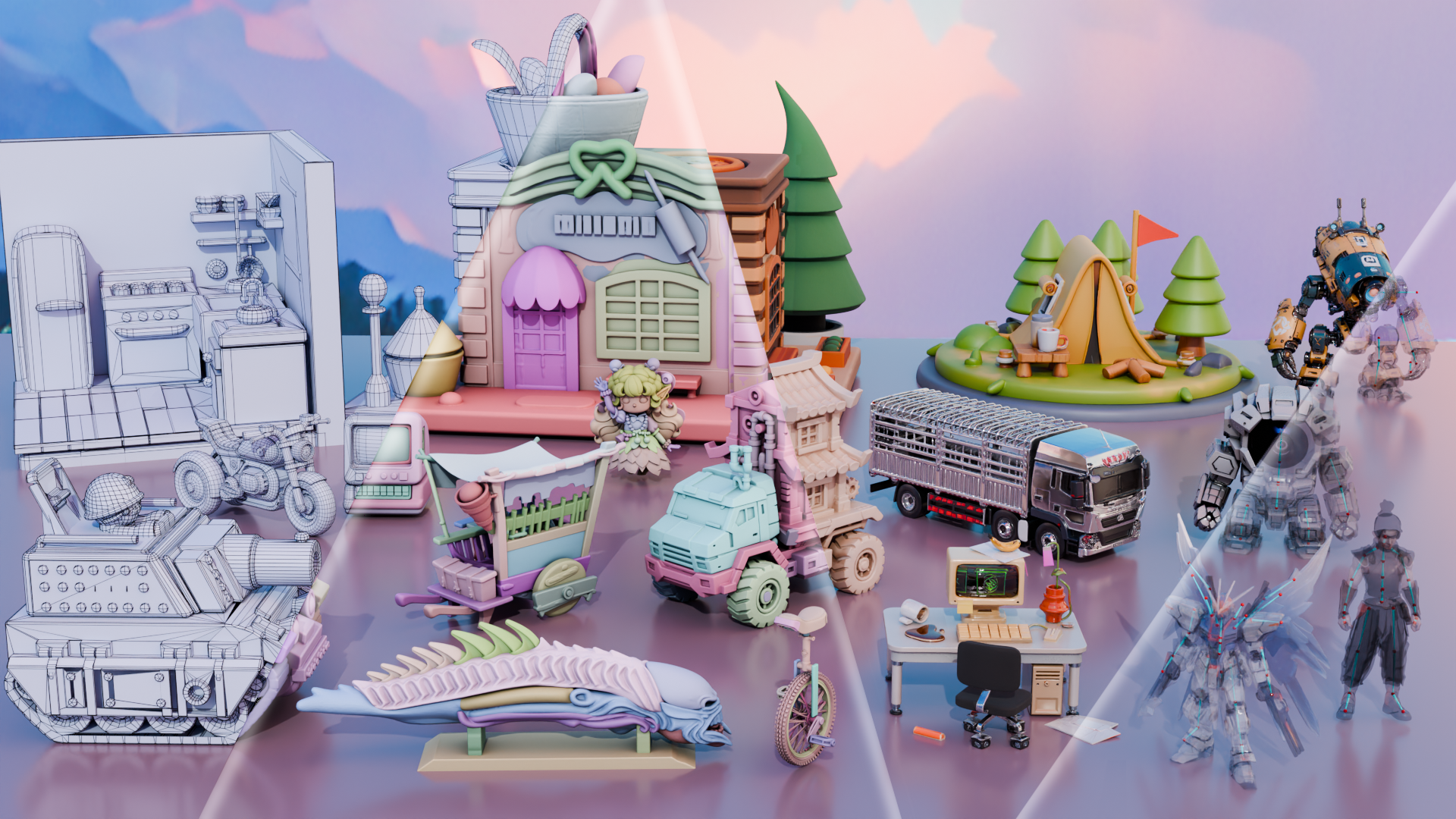}
    \caption{High quality 3D assets generated by Hunyuan3D Studio.}
    \label{fig:studio-teaser}
\end{figure}

\begin{abstract}
The creation of high-quality 3D assets, a cornerstone of modern game development, has long been characterized by labor-intensive and specialized workflows. This paper presents \textbf{Hunyuan3D Studio}, an end-to-end AI-powered content creation platform designed to revolutionize the game production pipeline by automating and streamlining the generation of game-ready 3D assets. At its core, Hunyuan3D Studio integrates a suite of advanced neural modules (such as Part-level 3D Generation, Polygon Generation, Semantic UV, etc.) into a cohesive and user-friendly system. This unified framework allows for the rapid transformation of a single concept image or textual description into a fully-realized, production-quality 3D model complete with optimized geometry and high-fidelity PBR textures. We demonstrate that assets generated by Hunyuan3D Studio are not only visually compelling but also adhere to the stringent technical requirements of contemporary game engines, significantly reducing iteration time and lowering the barrier to entry for 3D content creation. By providing a seamless bridge from creative intent to technical asset, Hunyuan3D Studio represents a significant leap forward for AI-assisted workflows in game development and interactive media.
\end{abstract}

\section{Introduction}

The demand for high-fidelity 3D content has surged, driven by the expanding frontiers of 3D games, virtual production, and the metaverse. However, traditional 3D asset creation remains a complex, time-consuming, and often costly endeavor, typically requiring expertise across multiple software suites for modeling, UV mapping, texturing, and rigging. This process can form a bottleneck in game production, limiting creative iteration and accessibility.

Recent advancements in generative AI~\citep{zhang2024clay, xiang2025structured, bpt, lei2024diffusiongan3d}, particularly diffusion models, have driven rapid progress in specific areas of 3D content creation, such as geometry generation where systems like Hunyuan3D series~\citep{yang2024hunyuan3d, zhao2025hunyuan3d, lai2025unleashing, hunyuan3d2025hunyuan3d, lai2025hunyuan3d} demonstrate scalable, high-resolution asset synthesis from single images or text prompts. However, despite these breakthroughs in shape formation, the field continues to struggle with integrating these advances into assets that simultaneously meet the dual demands of high visual fidelity and technical readiness for real-time rendering in game engines. Many existing solutions address only isolated parts of the pipeline (e.g., generating geometry without game-optimized topology or producing textures that lack material accuracy), leaving artists with the challenging task of integrating and refining these outputs into a usable, performant asset.

To bridge this critical gap, we introduce Hunyuan3D Studio, a comprehensive AI platform that reimagines the entire 3D asset creation workflow from the ground up. Our system is built upon the foundation of large-scale generative models but extends far beyond them into a fully integrated production environment. This report details the architecture of this end-to-end pipeline, which is designed to transform a high-level creative concept into a game-engine-ready asset with minimal manual intervention. We demonstrate that this integrated approach not only significantly accelerates content creation but also democratizes 3D artistry by lowering the technical barriers to producing production-quality assets.

This paper is organized as follows: Section ~\ref{sec:pipeline} provides a comprehensive overview of the Hunyuan3D Studio pipeline and its core modules. Sections ~\ref{sec:image} through ~\ref{sec:animation} delve into the technical specifics of each module. Finally, Section ~\ref{sec:conclusion} discusses conclusions, limitations, and future work.

\section{Hunyuan3D Studio Pipeline}  \label{sec:pipeline}

The Hunyuan3D Studio pipeline is architected as a sequential yet modular workflow, where each stage processes the asset and enriches it with data crucial for the next. This design ensures that the entire process—from an initial idea to a deployed game asset—is seamless, automated, and maintains the highest possible fidelity. The pipeline, as shown in Figure. \ref{fig:studio-pipeline}, comprises seven core technological modules, each addressing a fundamental stage in the asset creation process:

\begin{figure}[ht]
    \centering
    \includegraphics[width=1\linewidth]{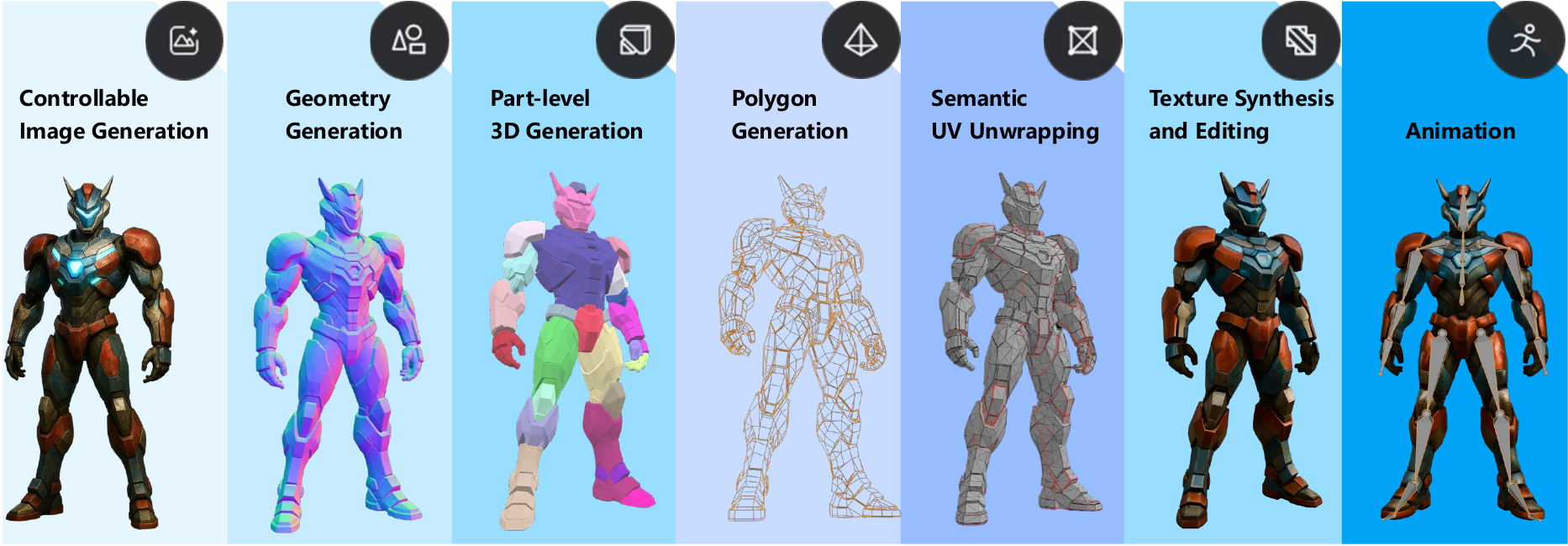}
    \caption{The pipeline of Hunyuan3D Studio.}
    \label{fig:studio-pipeline}
\end{figure}

\begin{itemize}
\item \textbf{Controllable Image Generation (Concept Design)}: The pipeline initiates with multi-modal input processing, supporting text-to-image and image-to-multi-view synthesis. A dedicated A-Pose standardization module ensures character models maintain consistent skeletal orientation, while neural style transfer adapts visual aesthetics to match target game art styles.

\item \textbf{High-Fidelity Geometry Generation}: This module generates detailed 3D geometry (high-poly mesh) from single or multi-view images, leveraging advanced diffusion-based architectures to ensure geometric alignment with input references and preserve intricate surface details.

\item \textbf{Part-level 3D Generation}: Using connectivity analysis and semantic partitioning algorithms, complex models are automatically decomposed into logical, functional components (e.g., a rifle's magazine, barrel, and stock), enabling independent editing and animation.

\item \textbf{Polygon Generation (PolyGen)}: This module abandons traditional graphics-based retopology methods and instead employs an autoregressive model for face-by-face generation to construct low-polygon assets. Taking a point cloud of the geometric surface as input, PolyGen intelligently retopologizes high-fidelity meshes, producing game-ready assets with low vertex counts and well-structured, deformation-aware edge flow.

\item \textbf{Semantic UV Unwrapping}: This module implements context-aware UV segmentation that groups surfaces by material type and texel density requirements, minimizing seams and ensuring efficient texture space utilization.

\item \textbf{Texture Synthesis and Editing}: Integrating generative models, the system produces physically-accurate PBR texture sets from text or image prompts, supported by a non-destructive editing layer for refinement via natural language commands.

\item \textbf{Animation Module}: The final automation stage infers joint placement and bone hierarchies, calculating vertex weights to create ready-to-animate assets that are configured for standard game engines.
\end{itemize}

These modules are orchestrated through a unified asset graph, where outputs from each stage propagate metadata to downstream processes. This enables parametric control, where high-level artistic adjustments cascade through the entire pipeline, and reversibility, allowing for incremental updates without full recomputation. The final output is configured and exported with all necessary specifications for the target game engine, such as Unity or Unreal Engine.

This modular yet integrated approach ensures Hunyuan3D Studio addresses the full spectrum of game asset creation—from conceptualization to engine integration—while maintaining artistic control and technical rigor. The following sections will provide a detailed technical dissection of each module's architecture and functionality.

\section{Controllable Image Generation}  \label{sec:image}
Our controllable image generation pipeline leverages state-of-the-art open-source models, comprising modules for image stylization and pose standardization, as described below.

\subsection{Image Stylization}

\begin{figure}[ht]
    \centering
    \includegraphics[width=0.85\linewidth]{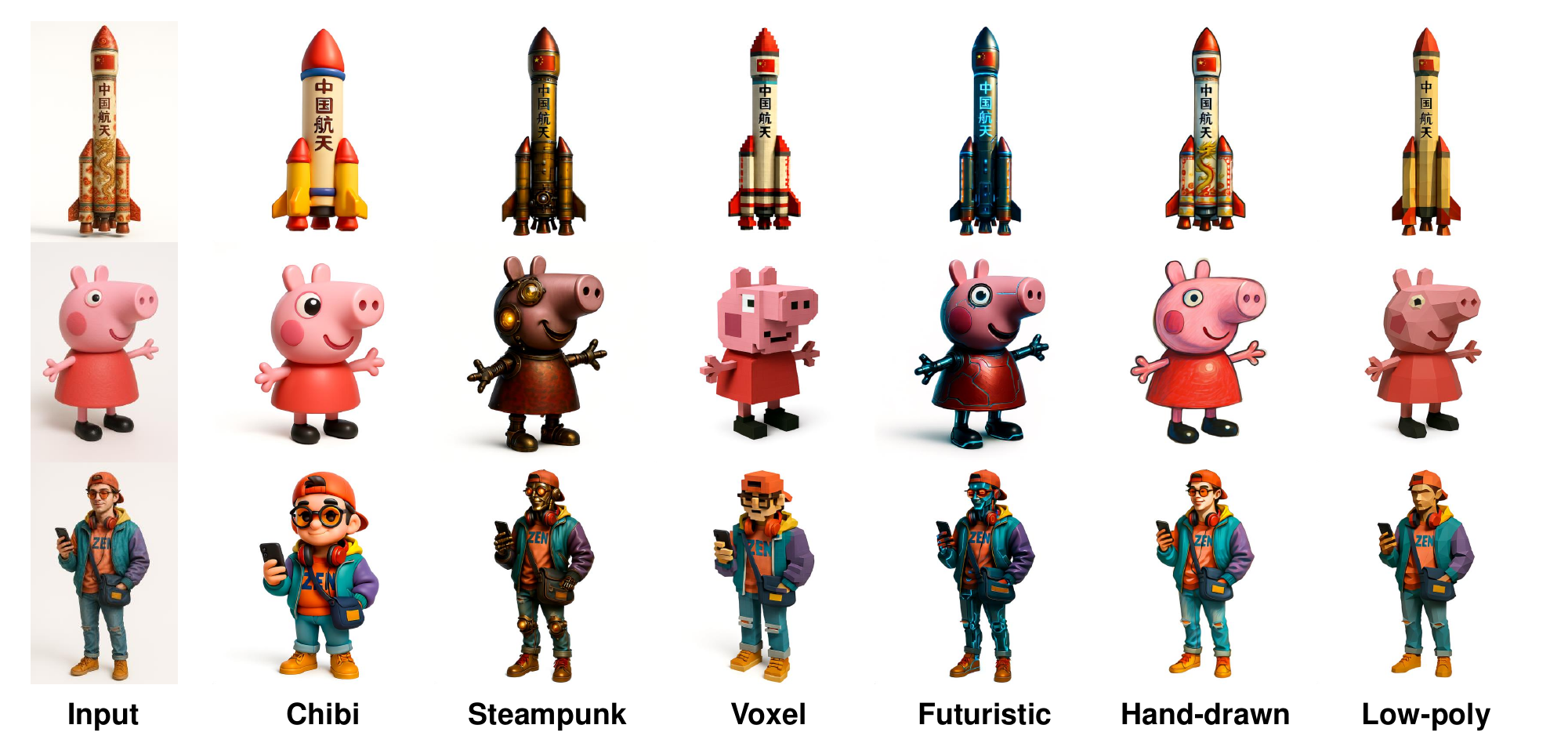}
    \caption{Visualization results of our image stylization module with pre-defined styles.}
    \label{fig:image-stylization}
\end{figure}

Our image stylization module enables users to generate 3D design drawings in diverse, pre-defined popular game art styles through a configurable option prior to 3D model generation, as shown in Figure. ~\ref{fig:image-stylization}. It employs a multi-style image-to-image generation model based on Qwen-Image-Edit \cite{wu2025qwen} and further adapted with Low-Rank Adaptation (LoRA) \cite{hu2022lora}. The model processes a user-provided subject image with a textual style instruction formatted as \textit{“Change the style to \{style type\} 3D model. White Background.”} to produce a stylized output that maintains content consistency with the input image while faithfully adhering to the specified artistic style.  The training data is constructed in a triplet format \textit{\{input reference image, style type, stylized 3D design drawing\}}, which establishes a precise correspondence between photorealistic subject images and their stylized counterparts. For text-to-image stylization, where no reference image is provided, the system first generates a reference image from the text prompt using an in-house general text-to-image model, and then processes it through the same image-to-image stylization pipeline to achieve the final stylized output.

\subsection{Pose Standardization}
\begin{figure}[ht]
    \centering
    \includegraphics[width=0.85\linewidth]{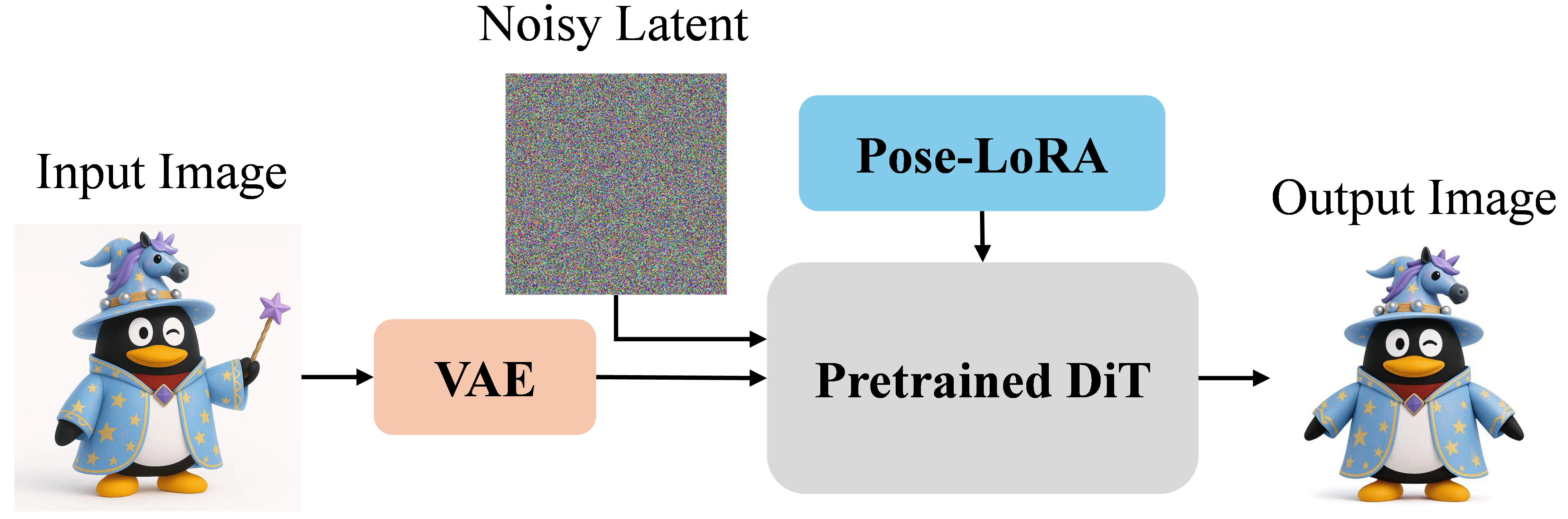}
    \caption{Overall workflow of our pose standardization module.}
    \label{fig:image-pose}
\end{figure}

Standardizing the pose (e.g., to an A-pose) from an arbitrary character reference image requires simultaneously achieving precise pose control and maintaining strict character consistency. In addition, it involves the removal of background elements and props from reference images. To achieve this, we leverage a character image with arbitrary poses/viewpoints as the reference and inject it as a conditioning input into FLUX.1-dev DiT to guide the generation process, as shown in Figure. \ref{fig:image-pose}. 

\begin{figure}[ht]
    \centering
    \includegraphics[width=0.85\linewidth]{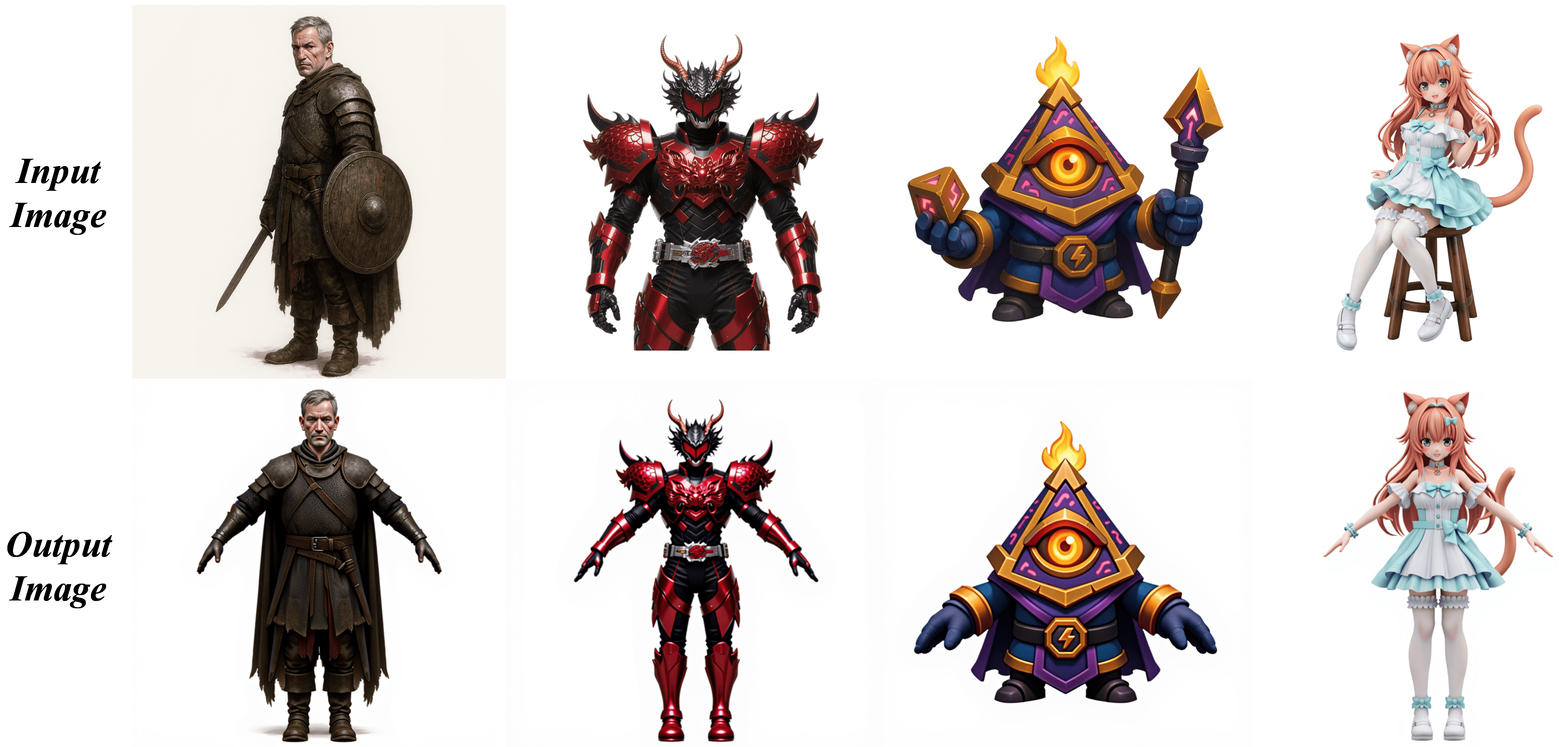}
    \caption{Visualization results of our pose standardization module.}
    \label{fig:image-pose-results}
\end{figure}

\begin{itemize}
  \item \textbf{Dataset Construction.} We first construct image pairs based on rendered character data in the following format: \textit{[character image with arbitrary poses/viewpoints, standard A-pose front views of the same character]}. Subsequently, rendered data that includes props (such as handheld weapons and pedestals) is processed through state-of-the-art editing models, including Flux-Kontext \cite{batifol2025flux}. This step isolates the character by removing all props and background elements, ensuring the subject's consistency is maintained. Finally, the resulting image pairs are manually curated and incorporated into the dataset to equip the model with the capability for prop and background removal.

  \item \textbf{Model Training.} We implement the progressive learning strategy, starting with an initial resolution of $512\times512$ pixels and increasing to $768\times768$. Consequently, the model learns to extract more intricate features, which substantially improves the fidelity of generated outputs, particularly in reproducing detailed facial characteristics and complex clothing textures. In addition, reference images of the same character under different scenarios are randomly selected and injected as conditional inputs, thereby achieving generalized pose control and consistent generation performance. Moreover, we have assembled supplementary high-quality datasets specifically targeting challenging categories,  including half-body portraits, non-human humanoids, and anthropomorphic characters. These datasets are used for post-training techniques like Supervised Fine-Tuning (SFT) and Direct Preference Optimization (DPO) to bolster the model's generalizability and robustness.
\end{itemize}

\section{High-Fidelity Geometry Generation}  \label{sec:geometry}

\subsection{Preliminary}
Our geometry-generation pipeline is built upon the state-of-the-art Hunyuan3D 2.1~\citep{hunyuan3d2025hunyuan3d} and Hunyuan3D 2.5~\citep{lai2025hunyuan3d} frameworks. The Hunyuan3D 2.1 framework comprises two modules:

\begin{itemize}
  \item \textbf{Hunyuan3D-ShapeVAE} — a variational encoder–decoder transformer that first compresses and then reconstructs 3-D geometry. The encoder receives a point cloud endowed with 3-D positions and surface normals $\{\mathbf{x}_i=(\mathbf{P}_i,\mathbf{N}_i)\mid\mathbf{P}_i,\mathbf{N}_i\in\mathbb{R}^3\}$, and embeds it into compact shape latents  $\mathbf{z}$ via a Vector-Set Transformer with importance sampling~\citep{zhang20233dshape2vecset}. The decoder employs $\mathbf{z}$ to query a 3-D neural field $\mathbf{F}_g\in\mathbb{R}^{D\times H\times W\times d}$ on a uniform grid $\mathbf{Q}_g\in\mathbb{R}^{D\times H\times W\times 3}$, and subsequently maps $\mathbf{F}_g$ to signed-distance values $\mathbf{S}_g\in\mathbb{R}^{D\times H\times W}$.

  \item \textbf{Hunyuan3D-DiT} — a flow-based diffusion model that operates in the latent space of ShapeVAE.  
  The network stacks 21 Transformer layers, each enhanced with a Mixture-of-Experts (MoE) sub-layer to significantly enlarge capacity and expressive power. Hunyuan3D-DiT is trained to map gaussian noises to shape latents $\mathbf{z}$ with the flow matching objective~\citep{lipman2022flow, esser2024scaling}.
\end{itemize}

\subsection{Conditional Generation}


\begin{figure}[ht]
  \centering
  \begin{subfigure}[b]{0.48\linewidth}
    \centering
    \includegraphics[width=0.95\linewidth]{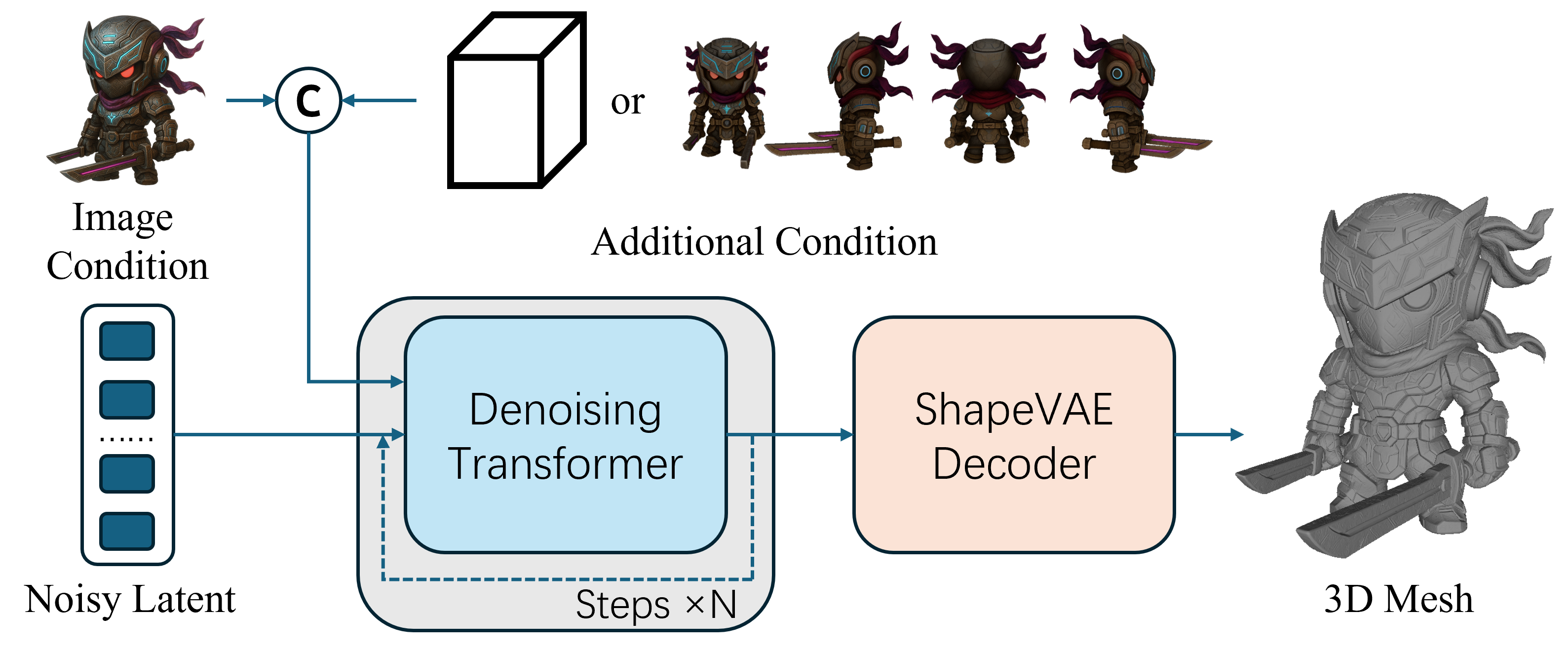}
    \caption{3D shape generation pipeline}
  \end{subfigure}
  \hfill
  \begin{subfigure}[b]{0.48\linewidth}
    \centering
    \includegraphics[width=0.95\linewidth]{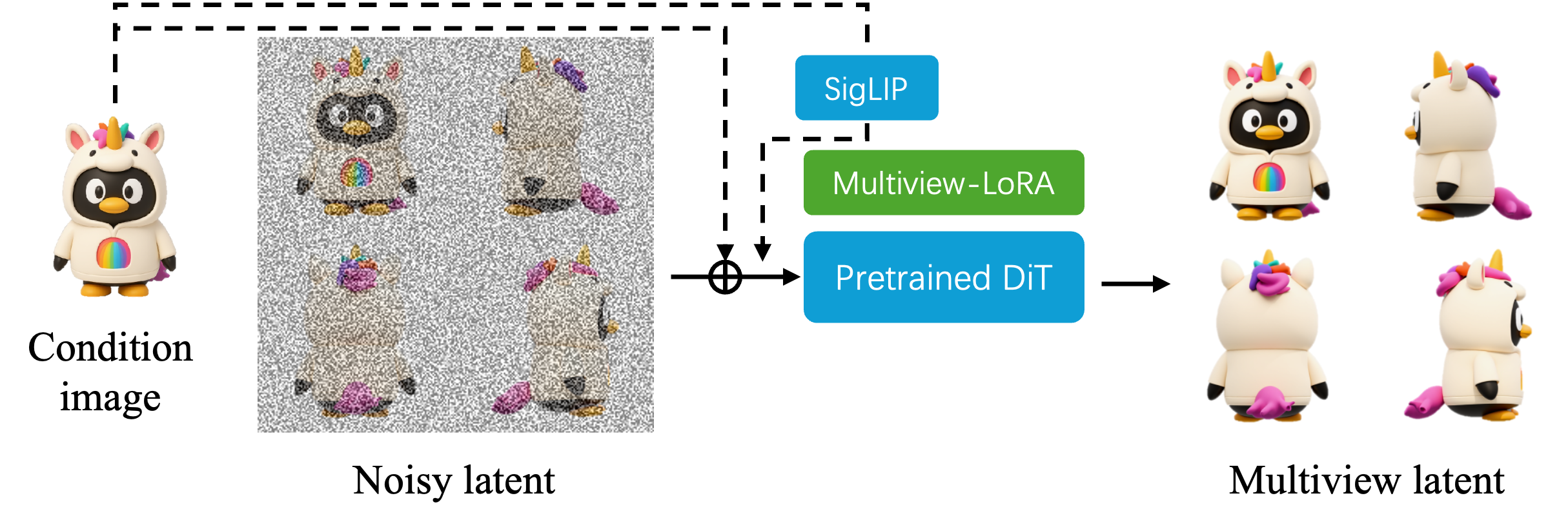}
    \caption{Image-to-multiview pipeline}
  \end{subfigure}
  \caption{Overview of conditional generation.}
  \label{fig:cond-gen pipe}
\end{figure}

Hunyuan3D-DiT is primarily conditioned on a single input image, which is first resized to 518 × 518, removed background and then encoded by a frozen DINOv2 backbone~\citep{oquab2023dinov2} to obtain the image latent $\mathbf{c}_I\in\mathbb{R}^{B\times L\times C}$, which is subsequently fused into the generated
shape latent via cross attention.

To provide additional geometric and prior guidance, Hunyuan3D-Studio supplies two additional control signals: an explicit 3-D bounding box and multi-view images. The overall pipeline is illustrated in Figure~\ref{fig:cond-gen pipe}(a).


\subsubsection{Bounding Box Condition}
To better align the generated 3D assets with user intent, we introduce a bounding box as the simplest 3-D control signal to steer the DiT model. Given a bounding box, we first encode its height, width and length into a single shape latent $\mathbf{c}_B\in\mathbb{R}^{B\times 1\times C}$ with a two-layer MLP. Then, $\mathbf{c}_B$ is concatenated with $\mathbf{c}_I$ along the sequence dimension to form the final conditioning vector.

During training we deliberately misalign the object proportions between the image and the point cloud—via mild deformation of either modality—to force the network to rely on the bounding-box signal.


\subsubsection{Generated Multi-view Image Condition}
To harness the powerful capabilities of image-generation models, we use multi-view images produced by an image diffusion model as the condition for character generation.

\textbf{Image to multiview image generation.} As, shown in Figure~\ref{fig:cond-gen pipe}(b), to synthesize high-fidelity multiview images from a single input, we introduce a lightweight module built upon a pretrained text-to-image foundation model~\citep{li2024hunyuandit}. This is achieved by training a Low-Rank Adaptation (LoRA) layer~\citep{hu2022lora}. The training process begins with the curation of a dataset comprising object-centric images from arbitrary camera poses, each paired with its corresponding ground-truth multiview images. During training, both the input single-view and target multiview images are encoded into their latent representations using the model's pretrained Variational Autoencoder (VAE). The LoRA layer is conditioned on two sources of information: (1) the noise-free latent of the single-view image, which is concatenated with the noised multiview latent, providing structural guidance; and (2) a semantic condition vector extracted from the input image using a pretrained SigLIP vision encoder~\citep{zhai2023siglip}. The LoRA parameters are then optimized using a standard flow-matching loss.

\textbf{Multi-view image injection.} Similar to the single image condition, we first encode all the images into image latents $\{\mathbf{c}_I^i|i={\rm org, front, left, back, right}\}$. Each non-original view is marked by a sinusoidal positional embedding with a fixed index. After positional encoding, the latents from generated views are concatenated with the original-image latent to form the final condition.


\subsection{Visualization}
\begin{figure}[ht]
    \centering
    \includegraphics[width=0.95\linewidth]{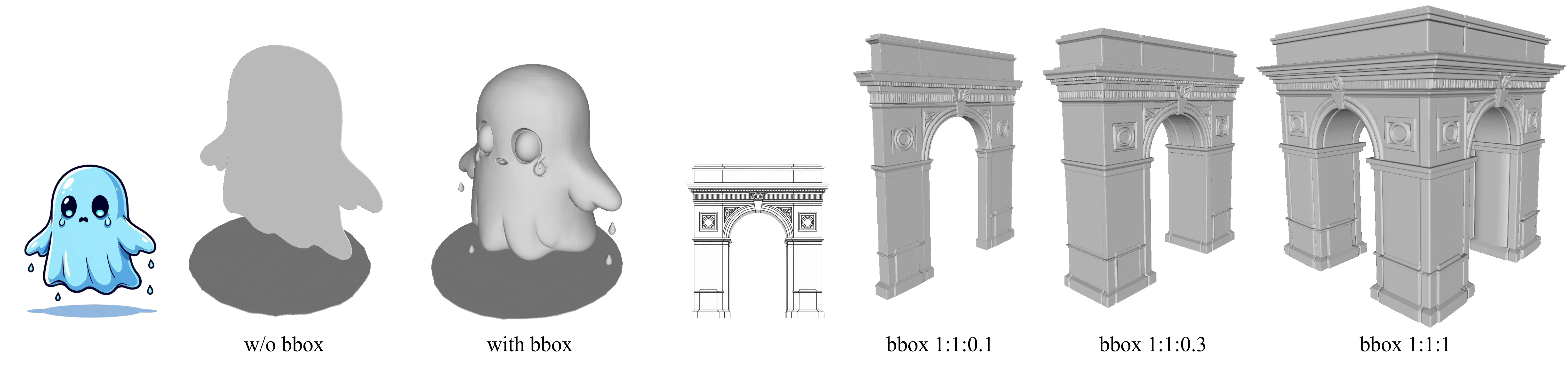}
    \caption{3D geometry generated with bounding box control.}
    \label{fig:image-generation bbox}
\end{figure}

\textbf{Bbox condition.} As illustrated in Figure~\ref{fig:image-generation bbox}, the bbox control signal not only succeeds in producing high-quality geometry when image-only geometric generation fails, but also generates 3D assets with appropriate proportions and well-structured forms according to the given bbox.

\begin{figure}[ht]
    \centering
    \includegraphics[width=0.95\linewidth]{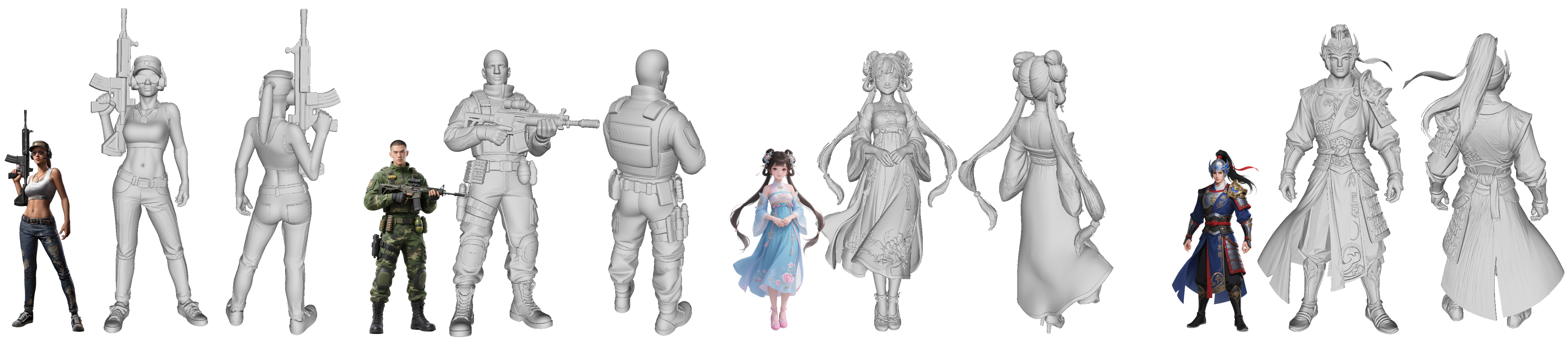}
    \caption{3D geometry generated with generated multi-view image control.}
    \label{fig:image-generation character}
\end{figure}

\textbf{Generated multi-view image condition.} Leveraging a state-of-the-art multi-view generation model as guidance, our approach produces high-fidelity 3D character assets, as exemplified in Figure~\ref{fig:image-generation character}.

\input{./sections/part}  \label{sec:part}

\input{./sections/polygen}  \label{sec:polygen}

\section{Semantic UV}  \label{sec:uv}

\begin{figure*}[htp]
    \centering
    \includegraphics[width=\textwidth]{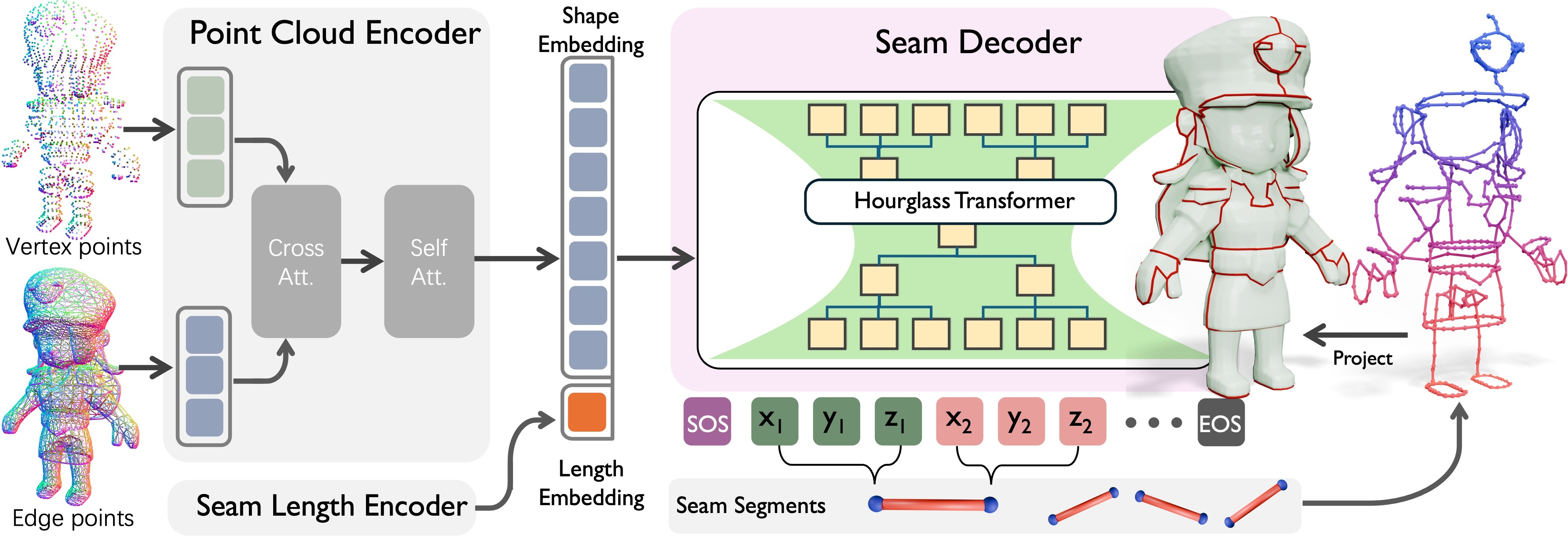}
    \caption{SeamGPT architecture:  Point cloud encoder extracts shape context; Causal transformer decoder generates axis-ordered seam coordinates. 
    Color indicates the prediction order is of the seam segments (red to blue).
    }
    \label{fig:pipeline}
\end{figure*}

The results of traditional UV unwrapping methods often lack semantic significance, which notably affects the quality of downstream texturing and the efficiency of resource utilization. Consequently, these traditional methods cannot be directly applied in professional pipelines, such as those used in game development and film production. To handle this challenge, we introduce SeamGPT, a novel framework that generates artist-style cutting seams through an auto-regressive approach. Our method formulates surface cutting as a sequence prediction problem, where cutting seams are represented as an ordered series of 3D line segments. Given an input mesh $\mathit{M}$, our goal is to generate seam edges $S = \{s^{i}\}_{i \in [N_s]}$. The overview of SeamGPT is shown in Fig.~\ref{fig:pipeline}. We first introduce our seam representation strategy in Sec.~\ref{sec.4.1}, which encodes cutting seams as sequential tokens. In Sec.~\ref{sec.4.2}, we detail our auto-regressive generation process, which mimics the sequential decision-making of professional artists. 

\subsection{Mesh Seam Representation}
\label{sec.4.1}
A seam sequence $S$ of $N_s$ segments $\{s^i\}_{i \in [N_s]}$ is defined as: $S = \{s^1, s^2, \ldots s^{N_s}\}$, where each segment $s^i$ is a 3D line segment represented by two vertices: $s^i = (p^i_h, p^i_t)$, i.e. head and tail. Each vertex $p$ is defined by its 3D coordinates: $p = (x, y, z)$. 
Thus, a seam sequence can be decomposed at multiple levels:
\begin{align}
 S &= \{s^1, s^2, \ldots s^{N_s}\} && \mathrm{Segment\quad level} \nonumber\\
 &= \{p^1_h, p^1_t, p^2_h, p^2_t, \ldots, p^{N_h}_t, p^{N_h}_t\} && \mathrm{Point \quad level} \label{eq:seam_levels} \\
 &= \{x^1_h, y^1_h, z^1_h, x^1_t, y^1_t, z^1_t, \ldots, x^{N_s}_t, y^{N_s}_t, z^{N_s}_t\} && \mathrm{Coord. \quad level} \nonumber
\end{align}
\textbf{Seam ordering.}
For an auto-regressive model to function properly, a consistent order of sequences is required. 
Following existing practice for mesh generation~\cite{siddiqui2023meshgpt, bpt, hao2024meshtron} and wireframe generation ~\cite{ma2024generating},
we first sort vertices $yzx$ order, where $y$ represents the vertical axis, and then sort two vertices within an edge lexicographically, placing the lowest $yzx$-ordered vertex first. 
Finally, seam edges are sorted in ascending $yzx$-order based on the sorted values of their vertices.
The resulting order can be seen through the color coding of the generated meshes presented in Figure~\ref{fig:pipeline}, i.e. from red to blue.

\textbf{Quantization of coordinates.} Autoregressive models typically sample from a multinomial distribution over a discrete set of possible values. To adhere to this convention, we quantize vertex coordinates into a fixed number of discrete bins. The quantization resolution—determined by the number of bins—directly affects the precision of the predicted seam. Higher quantization levels yield more detailed and accurate representations but also increase the complexity of the generation process. To balance precision and tractability, we employ 1024-level quantization, enabling effective representation of complex seams.

\subsection{Autoregressive Seam Prediction}
\label{sec.4.2}

In autoregressive seam prediction, a seam sequence $S$ is generated by sequentially predicting each coordinate $c_i$ based on its conditional probability given all previously generated coordinates $P(c_i | c_{<i})$. The probability of the entire seam is then given by the joint probability of all its coordinates:
\begin{equation}
    P(S) = \prod_{i=1}^{6N_s} P(c_i | c_{<i}).
    \label{eq:seam_arm}
\end{equation}
\textbf{Global Shape Conditioning.}
Point clouds are a flexible and universal 3D representation that can be efficiently derived from other 3D formats, including meshes.
We use a point cloud encoder to extract representative features for characterizing the input 3D shapes.
In the context of surface cutting, seams are encouraged to align with the vertices and edges of the original mesh, such that cutting the mesh along seams does not create excessive extra faces.
To guide the decoder in producing vertex and edge-aligned seam placement, instead of sampling point clouds uniformly, we sample structural points only on vertices and along edges.
Specifically, we sample a total of 61,440 points, evenly split between: 30,720 points on vertices and 30,720 points on edges.
If the input mesh has fewer than 30,720 vertices, we use repeated over-sampling.
Points along an edge are sampled uniformly by interpolating between its start and end points with K samples, where K is determined based on the edge's length.
Finally, the input points are fed into a jointly trained point cloud encoder from~\cite{hunyuan3d22025tencent}, which processes the point cloud through a series of cross- and self-attention layers and compresses the point cloud to a latent shape embedding of length 3072 and dimension 1024.
Another option to create shape embeddings is to use mesh encoders, such as~\cite{zhou2020fullymeshae}. However, the computational cost of mesh encoder does not scale well when the input has a large number of vertices. We show in the ablation study that point cloud conditioning produces much better results than mesh conditioning.

\textbf{Seam Count control.}
Given an input shape, multiple possible suitable cutting solutions exist.
Depending on the application requirements, one can make many cuts to decompose a mesh, or just a few.
To regulate the cutting granularity, we concatenate a length embedding to the shape embedding.
We find that modulating the length embedding directly controls cutting granularity.

\textbf{HourGlass Decoder Architecture.}
Following~\cite{hao2024meshtron}, we build an hourglass-like autoregressive decoder architecture to sequence at multiple levels of abstraction.
The architecture employs multiple Transformer stacks at each level, with transitions between levels managed by causality-preserving shortening and upsampling layers that bridge these hierarchical stages.
There are three hierarchical levels: coordinates, vertices, and edges.
The input coordinate sequence is shortened by a factor of 3 at the vertex level.
It is then further shortened by a factor of 2 at the edge level.
Both shortening and upsampling layers are implemented to preserve causality.
The expanded sequence is combined with higher-resolution sequences from earlier levels via residual connections, similar to U-Nets.

\textbf{Training Strategy.} We employ two loss functions to for model training: a cross-entropy loss for token prediction and a KL-divergence loss to regularize the shape embedding space, ensuring it remains compact and continuous. 
Training begins with a 2,000-step warm-up phase and is parallelized across 64 Nvidia H20 GPUs (98GB Mem.) with a total batch size of 128. 
The model converges after one week of training. During training, we first scale all samples to fit within a cubic bounding box in the range of $-1$ to $1$.
We then apply data augmentation techniques, including random scaling within $[0.95,1.05]$, random vertex jitter, and random rotation.

\begin{table}[!ht]
    \centering
 
    \resizebox{\textwidth}{!}{
    \renewcommand{\arraystretch}{1.5}
    \begin{tabular}{lccccccccccccc|c}
    \toprule
        
         ~ & Bimba & Lucy & Ogre & Armadi. & Bunny & Nefert. & Dragon & Homer & Happy & Fandi. & Spot & Arm & Cow & Avg. \\ \hline
        Xatalas~\cite{xatlas} & 15.44  & \textbf{0.01 } & \textbf{0.66}  & \textbf{0.17}  & 61.84  & \textbf{0.03}  & \textbf{0.22}  & \textbf{7.51}  & 99.84  & 8.41  & 12.77  & 29.98  & \textbf{1.94}  & 18.37  \\ \hline
        Nuvo~\cite{srinivasan2024nuvo} & 19.12  & 57.89  & 26.22  & 114.21  & 16.84  & 20.92  & 61.03  & 21.92  & 267.43  & 19.76  & 12.93  & 37.34  & 12.70  & 52.95  \\ \hline
        FAM~\cite{zhang2024flattenanything} & 12.10  & 35.14  & 11.55  & 59.87  & 7.33  & 11.21  & 904.89  & 14.19  & 23.00  & 12.21  & 9.37  & 20.98  & 8.49  & 86.95  \\ \hline
        Edge-CLS  & \textbf{8.14}   & 22.85   & 23.54   & 11.18   & \textbf{3.91}   & 4.67   & 15.39   & 20.16   & \textbf{27.37}   & 12.47   & \textbf{5.77}   & 87.20   & 9.20   & 19.37  \\ \hline
        Ours & 10.68  & \textbf{0.01}  & 2.01  & 2.47  & 50.47  & 0.12  & 0.56  & 10.28  & 61.68  & \textbf{8.15}  & 5.95  & \textbf{14.88}  & 2.24  & \textbf{13.04}  \\ \bottomrule

    \end{tabular}
    }
    \caption{Quantitative results on Flatten-Anything benchmark using the face distortion metric.}
    \label{fam-bench}

\end{table}

\begin{table}[!ht]
    \vspace{-10pt}
    \centering

    \resizebox{0.8\textwidth}{!}{
    \renewcommand{\arraystretch}{1.35}
    \begin{tabular}{lcccccccc|c}
    \toprule
    
&Bowl  & Ball  & Sheep  & Driver  & Chicken  & Apple  & Giraffe  & Bottle  & Avg.   \\ \hline

Xatalas~\cite{xatlas} &       0.91  &\textbf{ 0.26}  & \textbf{1.19}  & 4.61  & 2.36  & \textbf{3.11}  & 2.85  &\textbf{ 0.57}  & 1.98   \\ \hline
Nuvo~\cite{srinivasan2024nuvo}  &      3.99  & 1.33  & 10.43  & 33.07  & 9.79  & 15.39  & 21.04  & 6.02  & 12.63    \\ \hline
FAM~\cite{zhang2024flattenanything}   &     3.80  & 0.81  & 6.45  & 15.33  & 18.98  & 6.64  & 11.77  & 4.36  & 8.52    \\ \hline
Ours &    \textbf{0.49 }  & 0.31   & 1.39   &\textbf{ 4.25}   &\textbf{ 1.86}   & 4.02   & \textbf{2.59}   & 0.67   &\textbf{ 1.95}    \\ \bottomrule

    \end{tabular}
    }
    \caption{Quantitative results on Toys4K Benchmark using the face distortion metric.}
    \label{toys4k-bench}

\end{table}

\subsection{Experiment}

\begin{figure}[!t]
    \centering
    \includegraphics[width=0.85\textwidth]{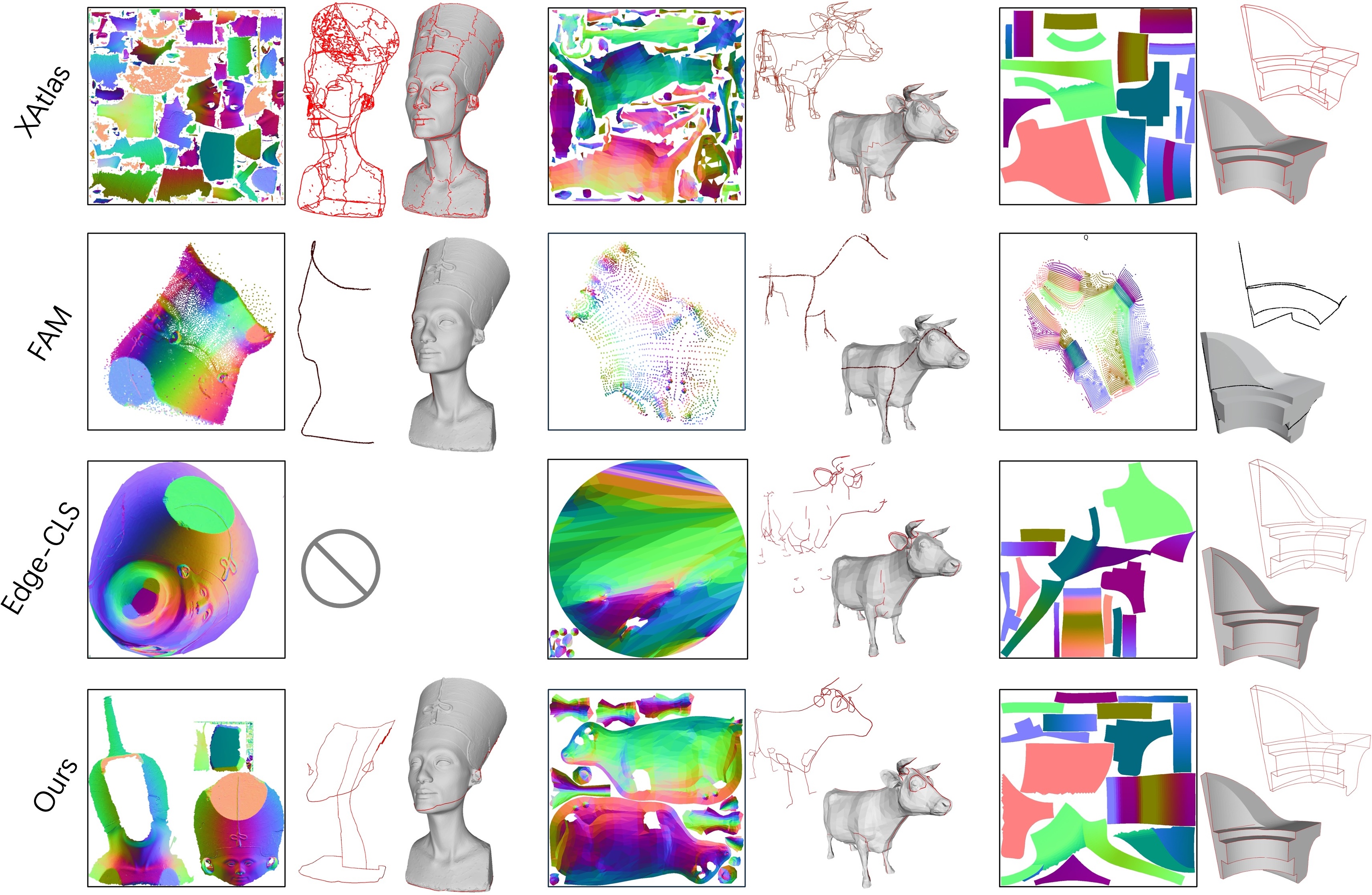}
\caption{Qualitative UV flatten results on FAM benchmark (  Nefertiti, Cow, and Fandisk). }
\label{fig:fam-compare}
\vspace{-10pt}
\end{figure}

\textbf{Benchmarks and Evaluation Metric.}
We conduct experiments on a diverse collection of 3D surface models from Flatten Anything (FAM)~\cite{zhang2024flattenanything}, which primarily includes low-poly meshes, CAD models, and 3D scanned meshes. We also evaluate on Toys4K~\cite{stojanov21cvpr}, a dataset of non-manifold artist-created meshes.  
Our evaluation leverages the \textbf{Mesh distortion} metrics, which is computed as the average conformal energy over all triangular faces of the mesh.

\textbf{Baselines and Implementations.} We compare SeamGPT against several state-of-the-art methods for mesh UV-unwrapping. \textbf{XAtlas}~\cite{xatlas} employs a bottom-up approach with bounded distortion charts. \textbf{Nuvo}~\cite{srinivasan2024nuvo} leverages neural fields with explicit parameterization constraints. \textbf{FAM}~\cite{zhang2024flattenanything} implements interpretable sub-networks in a bi-directional cycle mapping framework. 
We also built another baseline called \textbf{Edge-CLS}, which takes a mesh as input and uses graph convolution and Transformer layers to compute per-edge features. These features are then fed into an MLP classifier to predict whether each edge is a seam edge or not (i.e., this is an edge classification baseline).
We train Edge-CLS on the same training set and use the same UV-unwrapping process as SeamGPT.

\textbf{SeamGPT-based UV-unwrapping.} Once SeamGPT generates cutting seams, we implement a streamlined unwrapping process to create practical UV maps. We first map each predicted seam point to its nearest vertex on the input mesh, then connect these vertices through shortest geodesic paths along the mesh edges. We then cut the mesh by duplicating vertices along these paths, creating independent boundaries for flattening. Finally, we apply Blender's Minimum Stretch algorithm to the segmented mesh, optimizing UV coordinates to evenly distribute stretching while preserving the semantic structure defined by our seams. This process yields low-distortion UV mappings that respect functional and aesthetic boundaries, improving upon conventional automated methods.

\textbf{Comparison results.}
Tables~\ref{fam-bench} and~\ref{toys4k-bench}, along with Figure~\ref{fig:fam-compare}, present qualitative and quantitative results.
SeamGPT achieves the best performance across all metrics. In contrast: XAtlas generates over-fragmented cuts, FAM fails to produce subtle cuts consistently, Edge-CLS performs well only on sharp edge features but struggles with generating seams on smooth, featureless regions.
Our method consistently produces semantic and reasonable cuts regardless of surface characteristics.

\textbf{User study.} To further assess our method's practical utility, we conducted a user study with 20 professional 3D artists evaluating \textbf{Boundary} quality and \textbf{Editability}. Boundary quality measures how unfragmented a UV map is, while editability reflects how well the mapping supports appearance editing. Participants rated UV unwrappings from all methods on a 5-point scale. As shown in Table ~\ref{tab:userstudy}, SeamGPT significantly outperforms existing methods in both metrics.

\begin{table}[!ht]

\centering
\resizebox{\textwidth}{!}
{
\small
\renewcommand{\arraystretch}{1.2}

\begin{tabular}{l|cc|cc|cc|cc}
\toprule
\multicolumn{1}{c}{} & \multicolumn{2}{|c}{\texttt{Fandisk}} & \multicolumn{2}{|c}{ Cow} & \multicolumn{2}{|c}{\texttt{Nefertiti}} & \multicolumn{2}{|c}{ Avg. }\\
\multicolumn{1}{c|}{} & 
\multicolumn{1}{c}{Boundary $\uparrow$} & \multicolumn{1}{c|}{Editability $\uparrow$} & \multicolumn{1}{c}{Boundary $\uparrow$} & \multicolumn{1}{c|}{Editability $\uparrow$} & \multicolumn{1}{c}{Boundary $\uparrow$} & \multicolumn{1}{c|}{Editability $\uparrow$} & \multicolumn{1}{c}{Boundary $\uparrow$} & \multicolumn{1}{c}{Editability $\uparrow$} \\ \hline

Xatalas~\cite{xatlas}
& \textbf{4.42} & \textbf{4.42} & 2.68  & 2.37 & 2.79  & 2.47 & 3.30 & 3.09 \\ \hline

Nuvo~\cite{srinivasan2024nuvo}
& 1.32  & 1.32  & 1.16  & 1.21  & 1.42 & 1.42 & 1.30 & 1.32 \\ \hline

FAM~\cite{zhang2024flattenanything}
& 1.74  & 1.53 & 1.84  & 1.53 & 2.05  & 1.84 & 1.88 & 1.63 \\ \hline

Edge-CLS
& 3.89  & 3.84 & 2.79  & 2.32 & 2.58 & 2.16 & 3.09 & 2.77 \\ \hline

Ours
& 4.37  & 4.32 & \textbf{4.16}  & \textbf{4.16} & \textbf{3.47}  & \textbf{3.58} & \textbf{4.00} & \textbf{4.02} \\ 
\bottomrule
\end{tabular}
}
\caption{User Study about Boundary quality and Editability.}
\vspace{-10pt}
\label{tab:userstudy}
\end{table}

\subsection{Ablation Study}

\begin{wrapfigure}{r} {0.5\textwidth}
\vspace{-15pt}
\centering
\begin{center}
\includegraphics[width=\linewidth]{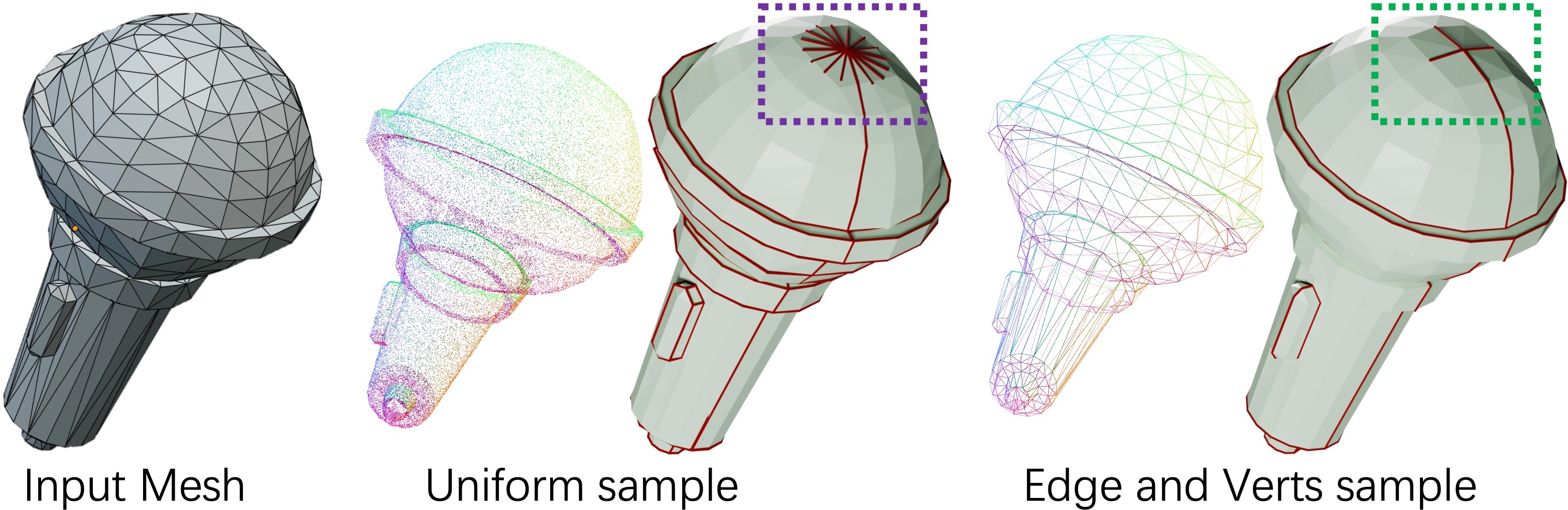}
\end{center}
\vspace{-5pt}
\caption{Ablation of point sampling strategy.}
\label{fig:ablation-point-sampling}
\vspace{-10pt}
\end{wrapfigure}

\textbf{Point cloud sampling strategy.} As shown in Figure.~\ref{fig:ablation-point-sampling}, when conditioned on point clouds uniformly sampled across the mesh surface, the generated seams remain logically valid from a surface-cutting perspective but may not precisely align with the input mesh's vertices and edges. 
In contrast, sampling point clouds along edges and vertices produces seams that naturally conform to the mesh topologies. This could prevents creating excessive extra mesh faces.
We also found that sampling along edges and vertices significantly improves model convergence, as the transformer gains explicit positional awareness of potential cutting coordinates.

\begin{wrapfigure}{r}{0.5\textwidth}  
\vspace{-5pt}
\centering
\begin{center}
\includegraphics[width=\linewidth]{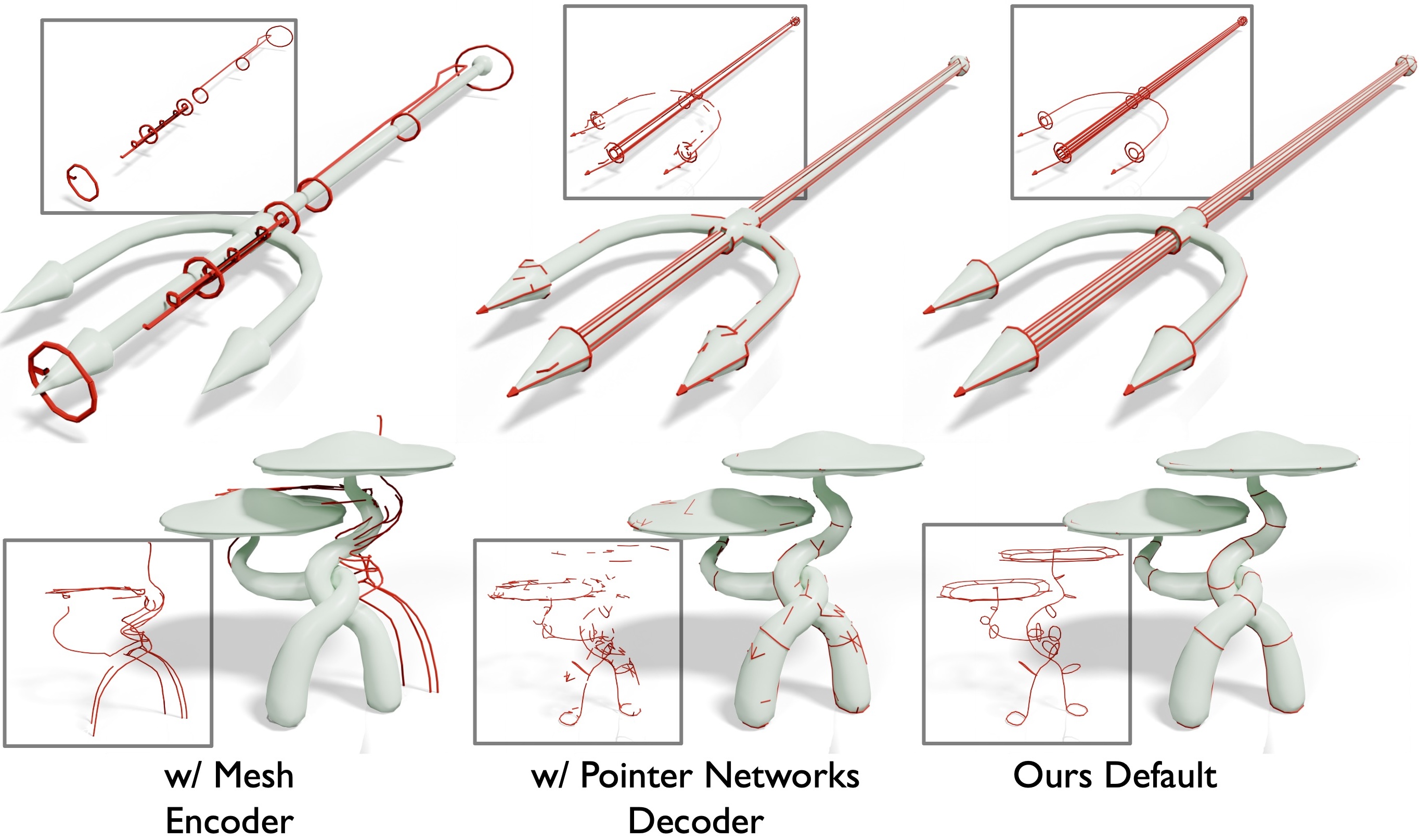}
\end{center}
\begin{minipage}{0.5\textwidth}  
\vspace{5pt}
    \caption{Ablation study of encoder and decoder.}
    \label{fig:ablation-ptrnet-meshcond}
\end{minipage}
\end{wrapfigure}

\textbf{Mesh encoder vs Point cloud encoder.}  An alternative approach for generating shape embeddings employs mesh encoders, as demonstrated by Zhou et al.~\cite{zhou2020fullymeshae}.
We implemented an encoder combining graph convolutions (operating on both vertices and edges) with a full self-attention transformer. This encoder produces vertex-wise tokens that are subsequently fed to the decoder via cross-attention mechanisms.
As shown in Figure~\ref{fig:ablation-ptrnet-meshcond}, point-cloud encoder yields superior results compared to mesh encoders. Furthermore, the computational cost of our mesh encoder scales poorly with increasing vertex counts.
Mesh encoder-based methods often fail to accurately capture the precise positions of original vertices, resulting in significant misalignment between the generated seam edges and the original mesh.

\textbf{Does Pointer networks works?}
In the case that the cutting seam forms a subset of the edges in the mesh, we can also adopt the Pointer Network~\cite{vinyals2015pointer-networks} architecture, which auto-regressively produce the pointers to the mesh edges.
We follow the implementation of Polygen~\cite{nash2020polygen} to build a pointer network with a mesh encoder that produces edge-wise embedding and a casual transformer to create pointers to the edges that lie on the seams auto-regressively. 
Pointer network struggles to generate consistent seams, often resulting in discontinuous cuts as demonstrated in Figure~\ref{fig:ablation-ptrnet-meshcond}.

\begin{figure}[!ht]
    \centering
    \includegraphics[width=1\textwidth]{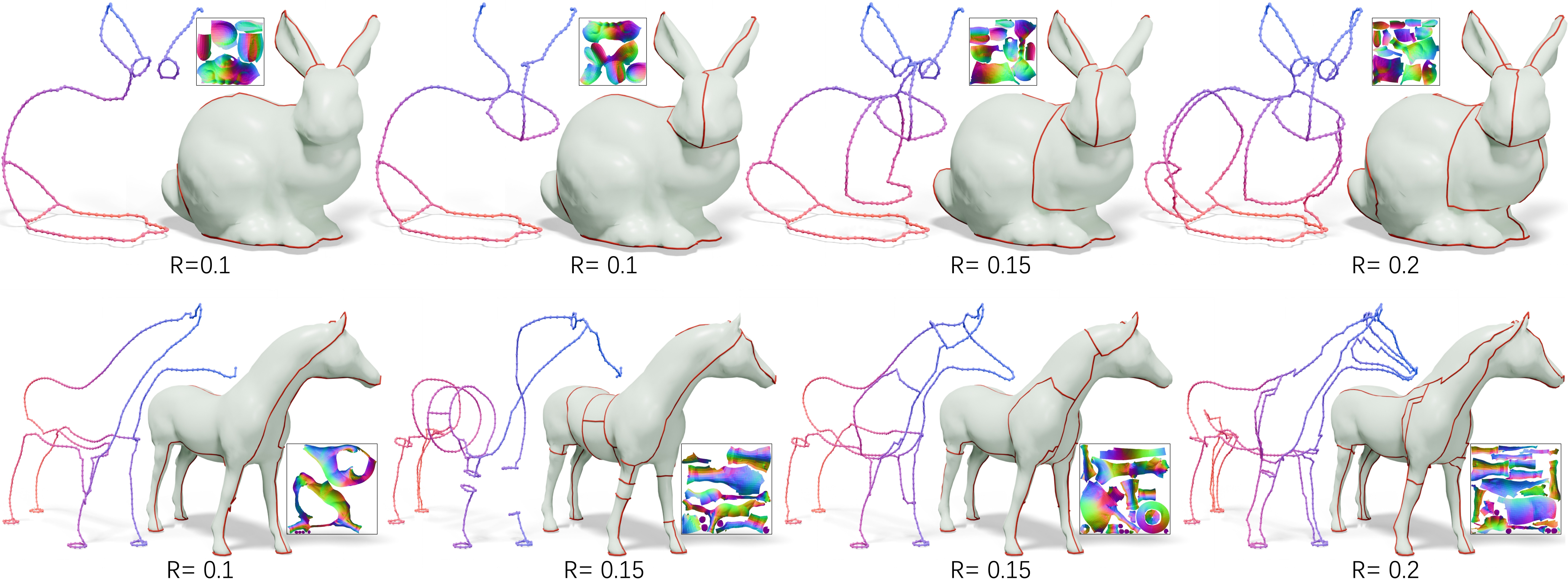}
\caption{Seam length control and diversity. We can control the cutting granularity by adjusting seam length. Diverse valid cutting seams can be generated.}
\label{fig:bunny-alation}
\end{figure}

\textbf{Seam length control and diversity.}  We define R as the ratio of seam segment count to the number of mesh vertices. Empirically, valid cutting seams typically has R value within the range [0.1,0.35]. Above this range result in over-cutting, while values below it lead to insufficient cuts. As shown in Figure~\ref{fig:bunny-alation}, controlling R allows us to adjust the granularity of the cuts.
Additionally, due to the non-deterministic nature of autoregressive transformers, we can generate diverse valid cutting seams from the same length control.

\section{Texture Generation and Editing}  \label{sec:texture}

Texture generation and editing technologies have critical importance in 3D asset creation. Physically Based Rendering (PBR) workflows rely on accurate texture maps to emulate real-world material behaviors under varying lighting conditions. High quality texture bridge the gap between geometric abstraction and perceptual realism, enhencing immersion and aesthetic coherence.

we introduced a high-fidelity texture synthesis methodology in ~\cite{zhao2025hunyuan3d, hunyuan3d2025hunyuan3d21imageshighfidelity, lai2025hunyuan3d} that lift a 2D diffusion model into a geometry-conditioned multi-view generative model i, subsequently baking its outputs into high-resolution texture maps via view projection. This framework systematically addresses two critical challenges in multi-view based texture generation: 
\begin{itemize}
\item Cross-view consistency and geometric alignment in ~\cite{feng2025romantexdecoupling3dawarerotary}.
\item Expansion of RGB textures into photorealistic PBR material textures in ~\cite{he2025materialmvpilluminationinvariantmaterialgeneration}.
\end{itemize}

In this report, we extend our texture generation framework into a comprehensive system supporting multimodal texture editing. First, we augment our existing multi-view PBR material generation model to accommodate text and image-guided multimodal editing. Second, we propose a material-based 3D segmentation method that generates part-wise material segmentation maps from input geometry-only meshes, enabling localized texture editing. Finally, we introduce a 4K material ball generation model that synthesizes high-resolution tileable texture balls—including Base Color, Metallic, Roughness, and Normal maps—from textual prompts, facilitating professional artistic workflows.

\begin{figure}[ht]
    \centering
    \includegraphics[width=1\textwidth]{figures/texture/text_image_edit.pdf
    }
    \caption{visualization results of multimodal texture editing: Perform text- and image-guided editing on textured meshes.}
    \label{fig:text_image_edit}
\end{figure}

\subsection{Multimodal Texture Editing}
We introduce a text-guided texture editing model trained on a meticulously curated dataset of 80k high-quality 3D assets with PBR materials. These assets were rendered into multi-view HDR images, and a Vision-Language Model (VLM) was employed to generate descriptive captions for textures and editing instructions. Leveraging the Flux Kontext framework, we constructed extensive image editing pairs across multiple viewpoints. Our texture foundation model then inferred consistent multi-view textures from these pairs, synthesizing a large-scale corpus of text-texture pairs for fine-tuning the editing model. During training, we followed flux kontext to unify textual prompts and reference image features into a joint latent sequence. Starting from a base texture generation model, the system was optimized end-to-end using 30,000 text-texture pairs, resulting in a unified model capable of texture synthesis and editing under both textual and visual guidance.

We introduce a streamlined Mixture of Experts (MoE) architecture for the image-guided texture editing model to handle diverse image inputs. To determine whether an input image aligns with the target geometry, we compute the CLIP similarity between geometrically rendered views and the input image. If the guidance image exhibits high geometric correspondence with the target mesh, we inject image features via a VAE encoder; otherwise, we use CLIP image embeddings for feature infusion—analogous to IP-Adapter's methodology. This adaptive conditioning mechanism ensures robust texture editing under arbitrary image conditions.

The fascinating multimodal editing results are demonstrated in Fig.~\ref{fig:text_image_edit}, indicating that we can perform material editing with diverse styles on objects in games, such as props and characters, both globally and locally.



\subsection{4K Matertial Map Generation}

\begin{figure}[ht]
    \centering
    \includegraphics[width=1\textwidth]{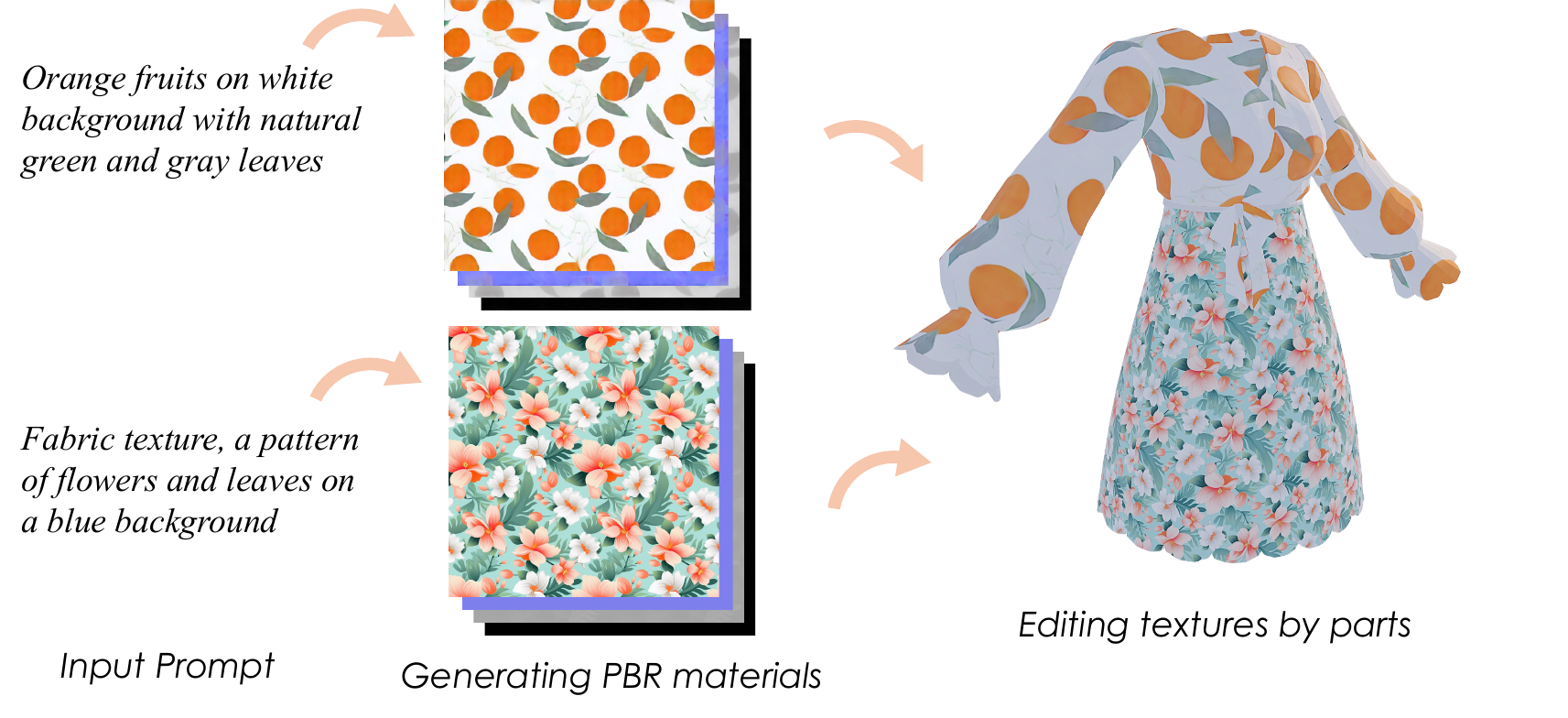}
    \caption{Visualization of material generation framework. }
    \label{fig:pbr_pipe}
\end{figure}

We innovatively adapt the 3D VAE framework~\ref{fig:pbr_pipe}—originally designed for encoding continuous video frames—to compress multi-domain material data (renders, base color, bump, roughness, metallic, etc.) into unified latent representations, enabling scalable 4K-resolution texture synthesis. Specifically, we fine-tune the 3D VAE using textured 3D assets to achieve domain-invariant feature extraction, resulting in a PBR-VAE module. Subsequently, we fine-tune a 3D Diffusion Transformer (DiT) with material ball datasets to establish the core architecture of our material ball generation model.

\section{Animation Module}  \label{sec:animation}

\subsection{Method overview}
In this section, we present the animation module, which is composed of two main branches: the humanoid character animation module and the general character animation module. Each character input is first processed by a detection module. If the input is identified as a humanoid character, it is directed to the humanoid animation branch; otherwise, it is routed to the general branch.

The humanoid branch consists of a template-based auto-rigging module and a motion retargeting module. To balance the accuracy of skeleton generation with ease of use, we adopt 22 body joints as the template skeleton. We follow~\cite{Guo_2025_CVPR} to construct our rigging and skinning model. However, unlike~\cite{Guo_2025_CVPR}, which does not incorporate rig-related information during skinning prediction, our model integrates both skeletal and vertex features to achieve more accurate results. In addition, our system includes a pose standardization module that converts user-provided models in arbitrary poses into a canonical T-pose. Feeding T-pose models into the motion retargeting module yields more reliable and precise outcomes.

In contrast, the general branch integrates an autoregressive skeleton generation module with a geometry topology-aware skinning module. Since general characters vary in both skeletal topology and the number of joints, most existing approaches to skeleton generation are based on autoregressive techniques, such as~\cite{Song_2025_CVPR,zhang2025unirig,guo2025auto,liu2025riganything}. These architectures have already demonstrated stability and accuracy in skeleton generation tasks, and our module is built upon these autoregressive methods. With respect to the skinning module, prior algorithms typically consider only mesh vertices and skeletal joints as input features, while paying little attention to the topological relationships among them. In contrast, our skinning module explicitly incorporates these topological relationships, leading to more robust and stable results.

\begin{figure}[t!]
\centering
  \begin{center}
    \includegraphics[width=1\textwidth]{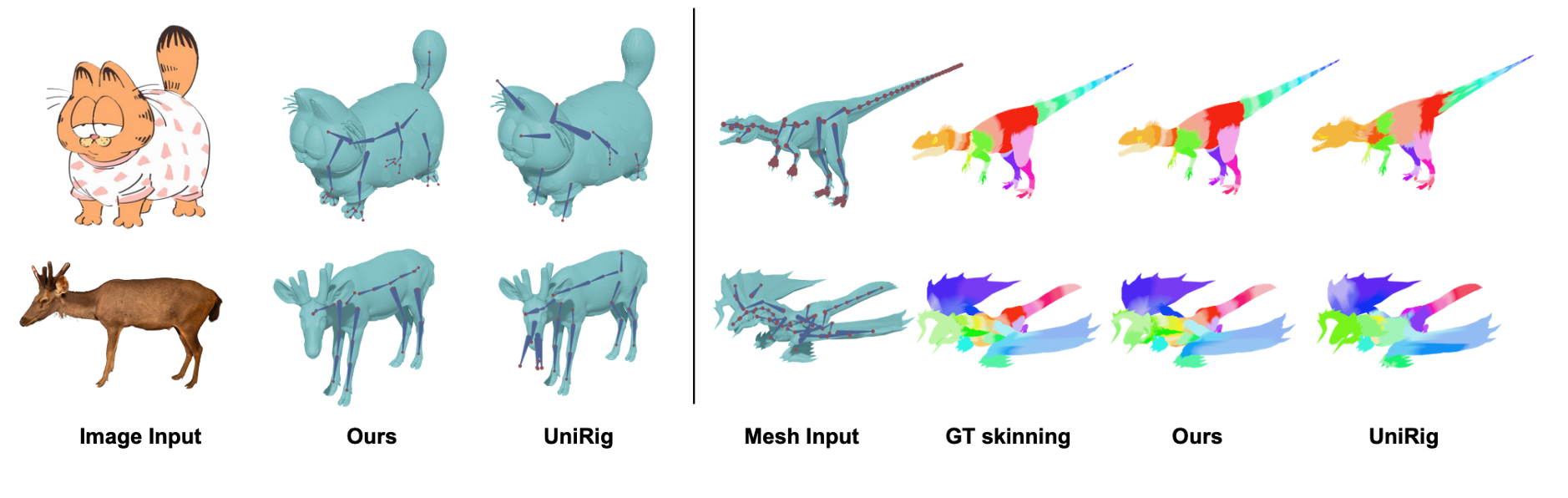}
  \end{center}
\caption{Comparison with UniRig~\cite{zhang2025unirig}. Left: results of rigging for a general character using our generated mesh. Right: results of skinning applied to the general character input.
}\label{fig:animation_result}
\vspace{-1em}
\end{figure}

\subsection{Implement details}
We used internally purchased and manually annotated datasets, which consist of approximately 80,000 high-quality general-character samples and 10,000 humanoid samples. All modules employ~\cite{zhao2023michelangelo} as the mesh encoder, while the autoregressive module is implemented using OPT-350M~\cite{zhang2022opt} as the transformer backbone. 

For the humanoid auto-rigging and skinning module, we apply motion-based data augmentation during training. The model is trained with a batch size of 6 on 8 H20 GPUs for 3 days. For the general-purpose rigging module, we train with a batch size of 16 on 24 H20 GPUs for 2 days. For the general-purpose skinning module, we adopt a batch size of 16 and train on 8 GPUs for 2 days.

\subsection{Qualitative results}

As shown in Figure~\ref{fig:animation_result}, our method produces more detailed results with fewer errors on general characters. Moreover, for the general skinning task, since our algorithm incorporates both skeletal and mesh topology information, it achieves higher overall accuracy compared to existing approaches.

\section{Conclusion}    \label{sec:conclusion}
\textbf{Hunyuan3D Studio} represents a paradigm shift in 3D content creation by integrating generative AI into a seamless, end-to-end pipeline that transforms single-image or text inputs into \textbf{game-ready assets} with optimized geometry, PBR textures, and engine-compatible topology. Our system’s core innovation lies in its modular yet unified architecture, which combines advanced neural models for geometry generation, component-aware segmentation, automated retopology (PolyGen), semantic UV unwrapping, and texture synthesis—all orchestrated to bridge the gap between creative intent and technical execution. Experimental validation confirms that assets generated by Hunyuan3D Studio meet the stringent requirements of modern game engines while significantly reducing production time and technical barriers. By democratizing access to high-quality 3D content, the platform empowers both artists and developers to iterate rapidly and focus on creativity rather than manual workflows.

\section{Contributors}

\begin{itemize}[leftmargin=0.25cm]
  \item \textbf{Project Sponsors:} Jie Jiang, Linus, Yuhong Liu, Di Wang, Tian Liu, Peng Chen
    \item \textbf{Project Leaders:}  Chunchao Guo, Zhuo Chen
    \item \textbf{Core Contributors:}
    \begin{itemize}[leftmargin=0.5cm]
    \item \textbf{PolyGen:} Biwen Lei, Jing Xu, Yiling Zhu, Haohan Weng, Jian Liu, Zhen Zhou, Jiankai Xing
    \item \textbf{Part:} Yang Li, Jiachen Xu,  Changfeng Ma, Xinhao Yan, Yunhan Yang, Chunshi Wang
    \item \textbf{UV:} Xinhai Liu, Duoteng Xu, Xueqi Ma, Yuguang Chen, Jing Li
    \item \textbf{Texture:} Shuhui yang, Mingxin Yang, Sheng Zhang, Yifei feng, Xin Huang, Di Luo, Zebin He
    \item \textbf{Animation:} Lixin Xu, Puhua Jiang, Changrong Hu, Zihan Qin, Shiwei Miao
    \item \textbf{Geometry:} Jingwei Huang, Haolin Liu, Yunfei Zhao, Zeqiang Lai, Qingxiang Lin, Zibo Zhao, Kunhong Li, Huiwen Shi
    \item \textbf{Image:} Ruining Tang, Xianghui Yang, Xin Yang, Yuxuan Wang, Zebin Yao 
    \end{itemize}
      
    \item \textbf{Contributors:}   
    \begin{itemize}[leftmargin=0.5cm]
    \item \textbf{Engineering:} Yihang Lian, Sicong Liu, Xintong Han, Wangchen Qin, Caisheng Ouyang, Jianyin Liu, Tianwen Yuan, Shuai Jiang, Hong Duan, Yanqi Niu, Wencong Lin, Yifu Sun, Shirui Huang, Lin Niu, Gu Gong, Guojian Xiao, Bojian Zheng, Xiang Yuan, Qi Chen, Jie Xiao, Dongyang Zheng, Xiaofeng Yang, Kai Liu, Jianchen Zhu
    \item \textbf{Data:} Lifu Wang, Qinglin Lu, Jie Liu, Liang Dong, Fan Jiang, Ruibin Chen, Lei Wang, Chao Zhang, Jiaxin Lin, Hao Zhang, Zheng Ye, Peng He, Runzhou Wu, Yinhe Wu, Jiayao Du, Jupeng Chen 
   \item \textbf{Art Designer:} Xinyue Mao, Dongyuan Guo, Yixuan Tang, Yulin Tsai, Yonghao Tan, Jiaao Yu, Junlin Yu, Keren Zhang, Yifan Li
   
    \end{itemize}
\end{itemize}

\clearpage

\bibliography{colm2024_conference}
\bibliographystyle{colm2024_conference}

\end{document}

%% file: math_commands.tex
\usepackage{amsmath,amsfonts,bm}

\def\eqref#1{equation~\ref{#1}}

\def\1{\bm{1}}

\DeclareMathAlphabet{\mathsfit}{\encodingdefault}{\sfdefault}{m}{sl}
\SetMathAlphabet{\mathsfit}{bold}{\encodingdefault}{\sfdefault}{bx}{n}

%% file: sections/part.tex
\section{Part-level 3D Generation}

\begin{figure}[!ht]
    \centering
    \includegraphics[width=1\linewidth]{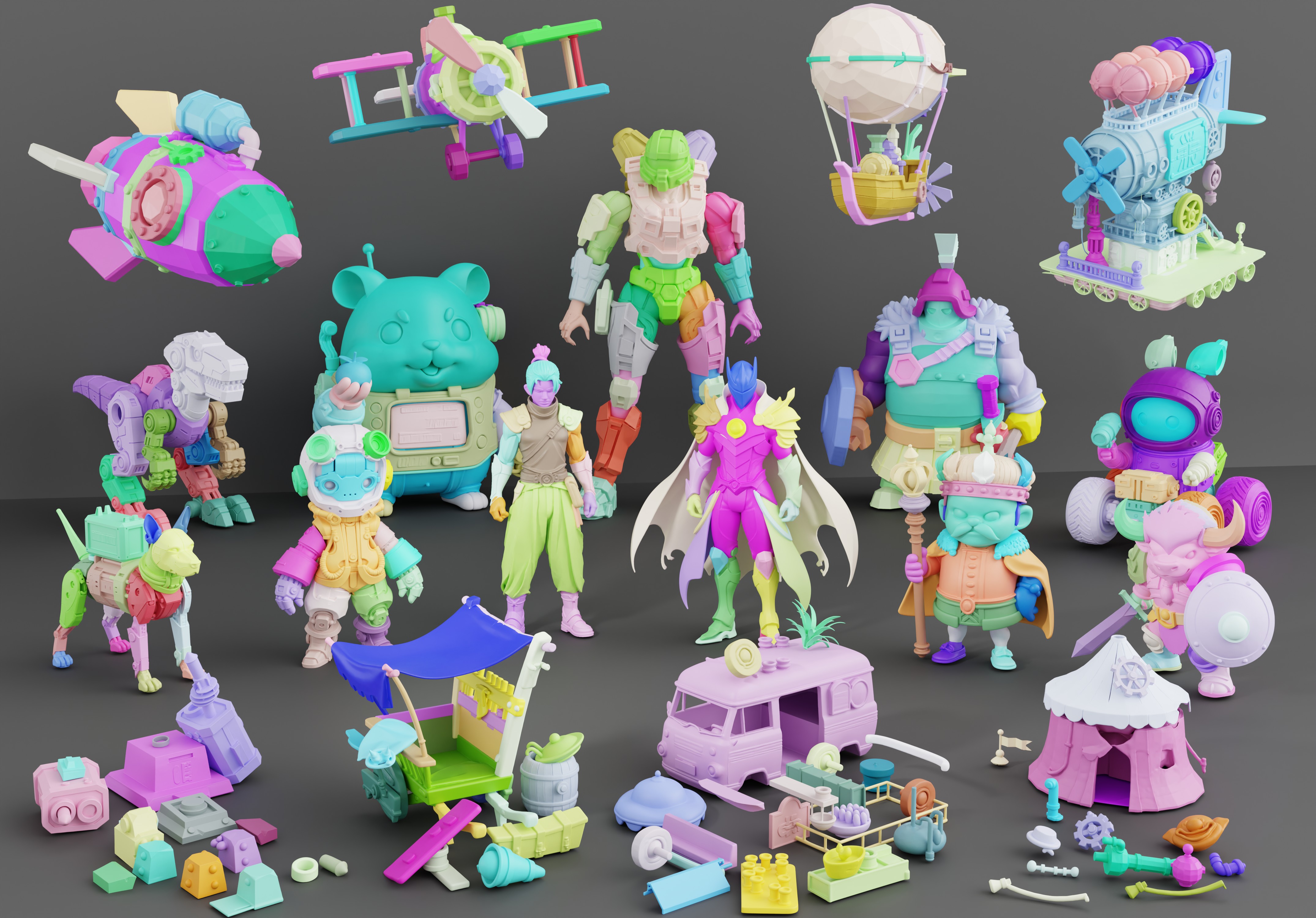}
    \caption{Our part level shape generation results.}
    \label{fig:part-teaser}
\end{figure}

Generating 3D shapes at part level is pivotal for downstream applications such as mesh retopology, UV mapping, and 3D printing.
However, existing part-based generation methods often lack sufficient controllability, produce inadequate geometric quality in generated parts, and suffer from limited semantic coherence.
This section establishes a new paradigm for creating production-ready, editable, and structurally sound 3D assets. Fig.~\ref{fig:part-teaser} shows our part-level shape generation results.

\begin{figure}[!ht]
    \centering
    \includegraphics[width=1\linewidth]{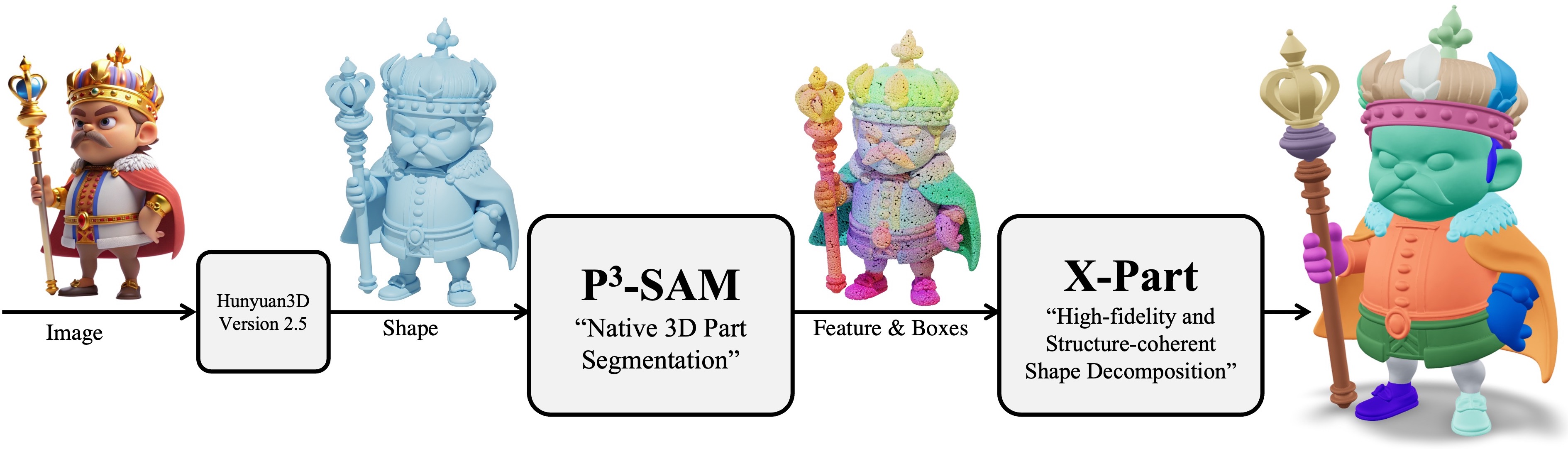}
    \caption{Pipeline of our image to 3D part generation. Given an input image, we first obtain the holistic shape using Huyuan3D 2.5~\cite{lai2025hunyuan3d}. 
The holistic mesh is then fed to part detection module \textbf{P$^3$-SAM}~\cite{P3_SAM} to obtain the semantic features and part bounding boxes.}
    \label{fig:part-whole-pipeline}
\end{figure}

As shown in Figure. ~\ref{fig:part-whole-pipeline}, given an input image, we first obtain the holistic shape using Huyuan3D 2.5~\cite{lai2025hunyuan3d}. 
The holistic mesh is then fed to part detection module \textbf{P$^3$-SAM}~\cite{P3_SAM} to obtain the semantic features and part bounding boxes.
Finally, \textbf{X-Part}~\cite{X-part} decompose the holistic shape into parts.
\textbf{P$^3$-SAM} and \textbf{X-Part} will be introduced in section~\ref{sec:p3sam} and~\ref{sec:x-part}

\begin{figure}[!t]
    \centering
    \includegraphics[width=1\linewidth]{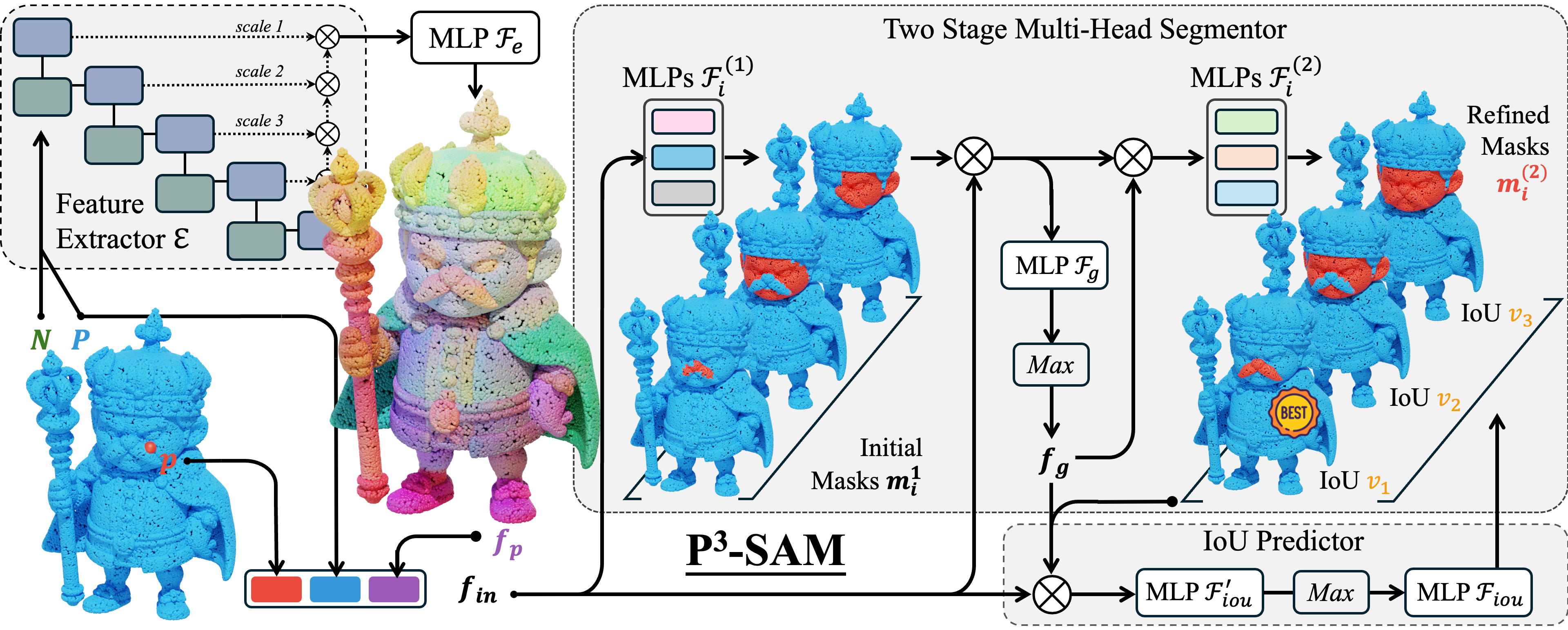}
    \caption{Training pipeline of \textbf{P}$^3$-SAM.}
    \label{fig:p3sam-pipeline}
\end{figure}

\subsection{ P$^3$-SAM: Native 3D Part Segmentation~\cite{P3_SAM}  }\label{sec:p3sam}

3D part segmentation is a fundamental step in our part generation pipeline.
In this section, we propose a native 3D \underline{\textbf{P}}oint-\underline{\textbf{P}}romptable \underline{\textbf{P}}art segmentation model termed \underline{\textbf{P}}$^3$-SAM, designed to fully automate the segmentation of any complex 3D objects into components with precise mask and strong robustness.
As a pioneering promptable image segmentation work, SAM provides a feasible implementation approach. 
However, our method focuses on achieving precise part segmentation automatically, and we simplify the architecture of SAM.
Without adopting the complex segmentation decoder and multiple types of prompts from SAM, our model is designed to handle only one positive point prompt.

Specifically, as shown in Fig.~\ref{fig:p3sam-pipeline}, \textbf{P}$^3$-SAM contains a feature extractor, three segmentation heads, and an IoU prediction head.
We employ PointTransformerV3 as our feature extractor and integrate its features from different levels as extracted point-wise features.
The input point prompt and feature are fused and passed to the segmentation heads to predict three multi-scale masks and an IoU (Intersection of Union) predictor is utilized to evaluate the quality of the masks.

\begin{figure}[!t]
    \centering
    \includegraphics[width=1\linewidth]{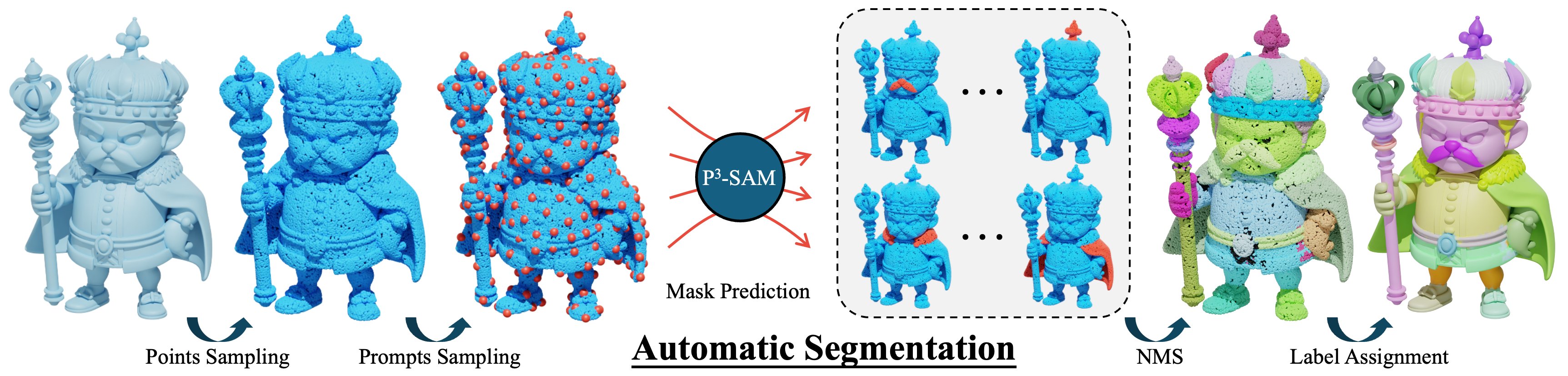}
    \caption{Pipeline of automatic segmentation using \textbf{P}$^3$-SAM.}
    \label{fig:p3sam-seg_pipe}
\end{figure}

To automatically segment an object, as shown in Fig.~\ref{fig:p3sam-seg_pipe}, we apply our segmentation model using point prompts sampled by FPS (Farthest Point Sampling) and utilize NMS (Non-Maximum Suppression) to merge redundant masks.
The point-level masks are then projected onto mesh faces to obtain the part segmentation results.

Another key aspect of this method is to eliminate the influence of 2D SAM, and rely exclusively on raw 3D part supervision for training a native 3D segmentation model.
While existing 3D part segmentation datasets are either too small  or lack part annotation, this work addresses the data scarcity by developing an automated part annotation pipeline for artist-created meshes and used it to generate a dataset comprising 3.7 million meshes with high-quality part-level masks.
Our model demonstrates excellent scalability with this dataset and achieves robust, precise, and globally coherent part segmentation. 

For more details of \textbf{P}$^3$-SAM, please refer to the paper~\cite{P3_SAM}. 

\paragraph{Comparison with SOTA.}
We evaluate each method on three datasets: PartObj-Tiny, PartObj-Tiny-WT, and PartNetE. 
PartObj-Tiny is a subset of Objarvse, containing 200 data samples across 8 categories, with manually annotated part segmentation information.
PartObj-Tiny-WT is the watertight version of PartObj-Tiny. To evaluate the performance of various networks on watertight data, we converted the meshes from PartObj-Tiny to watertight versions and successfully obtained 189 watertight meshes. 
PartNetE, derived from PartNet-Mobility, contains 1,906 shapes covering 45 object categories in the form of point clouds. We also evaluate various networks on it to verify their generalization performance on point cloud.
Table \ref{tab:p3sam-main_compare} and \ref{tab:main_compare_wt}  confirming the  superior performance of \textbf{P}$^3$-SAM under diverse conditions.

\begin{table}
\small
\centering
\setlength{\tabcolsep}{5pt}
\begin{tabular}{c|r|cccccccc|c} 
\toprule
Task       & Method    & Human & Animals & Daily & Build. & Trans. & Plants & Food  & Elec.  & AVG.    \\ 
\midrule
          & Find3D    & 23.99 & 23.99   & 22.67 & 16.03    & 14.11  & 21.77  & 25.71 & 19.83 & 21.28    \\
Fully   & SAMPart3D & 55.03 & 57.98   & 49.17 & 40.36    & 47.38  & 62.14  & \textbf{64.59} & 51.15 & 53.47     \\
Seg. w/o  & SAMesh    & \textbf{66.03} & \textbf{60.89}   & 56.53 & 41.03    & 46.89  & 65.12  & 60.56 & 57.81 & 56.86     \\
 Connect. & PartField & 54.52 & 58.07   & 56.46 & 42.47    & 49.09  & 59.16  & 55.4  & 56.29 & 53.93    \\
          & Ours      & 60.77 & 59.43   & \textbf{62.98} & \textbf{50.82}    & \textbf{57.72}  & \textbf{70.53}  & 54.04 & \textbf{61.96} & \textbf{59.88}    \\ 
\midrule
Seg. w/     & PartField & 80.85 & 83.43   & 77.83 & \textbf{69.66}    & \textbf{73.85}  & 80.21  & 85.27 & \textbf{82.30}  & 79.18  \\
Connect.    & Ours      & \textbf{80.77} & \textbf{86.46}   & \textbf{80.97} & 67.77    & 68.44  & \textbf{90.30}   & \textbf{92.90}  & 81.52 & \textbf{81.14}   \\ 
\midrule
\multirow{2}{*}{Interact.}          & Point-SAM & 8.63  & 9.38    & 17.47 & 11.19    & 7.63   & 13.95  & 23.02 & 12.73 & 13.00    \\
                                     & Ours      & \textbf{49.01} & \textbf{53.45}   & \textbf{52.36} & \textbf{38.50}     & \textbf{51.52}  & \textbf{62.57}  & \textbf{50.80}  & \textbf{51.86} & \textbf{51.23}   \\
\bottomrule
\end{tabular}
\caption{The comparison of our method with previous methods on PartObjectarverse-Tiny. The first two blocks represent class-agnostic part segmentation without and with connectivity, respectively, and the last block represents interactive segmentation.}
\label{tab:p3sam-main_compare}
\end{table}

\begin{table}[!ht]
\small
\centering
\begin{tabular}{r|ccccc|cc} 
\toprule
Task     & \multicolumn{5}{c|}{Fully Segmentation w/o Connectivity}               & \multicolumn{2}{c}{Interactive Seg.}  \\ 
\midrule
Method   & Find3D & SAMPart3D & SAMesh & PartField & Ours  & Point-SAM & Ours                      \\ 
\midrule
PartObj-Tiny-WT & wait   & wait      & wait   & wait      & \textbf{55.35} & 13.11     & \textbf{49.11}                     \\
PartNetE & 21.69   &  56.17    & 26.66   &  59.1     & \textbf{65.39} & 15.06     & \textbf{63.48}                     \\
\bottomrule
\end{tabular}
\caption{The comparison of our method with previous methods on the watertight version of PartObjectarverse-Tiny.}
\label{tab:main_compare_wt}

\end{table}

\subsection{ $\mathcal{X}$-Part: high-fidelity and structure-coherent shape decomposition ~\cite{X-part} }\label{sec:x-part}

This section shows how to decompose the shapes into parts.
Decomposing a complete 3D shape into meaningful semantic parts would greatly facilitate various downstream tasks. 
For instance, breaking down a complex geometry into simpler parts can significantly ease the process of mesh re-topology  and uv-unwrapping.
However, generating shapes at the part level presents two major challenges: 1) The decomposed geometry must maintain meaningful part-level semantics, and 2) The generation process must recover geometrically plausible structures for internal regions.

Mainstream part-generation methods adopt the latent vecset diffusion framework~\cite{zhao2025hunyuan3d}, where each part is represented as an independent set of latent codes for diffusion. 

The generation process can be executed independently for individual parts (e.g., HoloPart~\cite{yang2025holopart}) or simultaneously for all parts (e.g., PartCrafter~\cite{lin2025partcrafter}, PartPacker~\cite{tang2024partpacker}) to enhance part synchronization. Furthermore, multi-view or 3D segmentation are frequently employed for better part decomposition~\cite{yang2025holopart,yang2025omnipart}. However, these approaches are highly sensitive to inaccuracies in the segmentation results. Alternative works~\cite{lin2025partcrafter,tang2024partpacker} do not explicitly rely on segmentation, but they still fail to offer controllable part-based generation and often produce decomposed parts with ambiguous boundary.

\begin{figure*}[!t]
  \centering
  \includegraphics[width=1\linewidth]{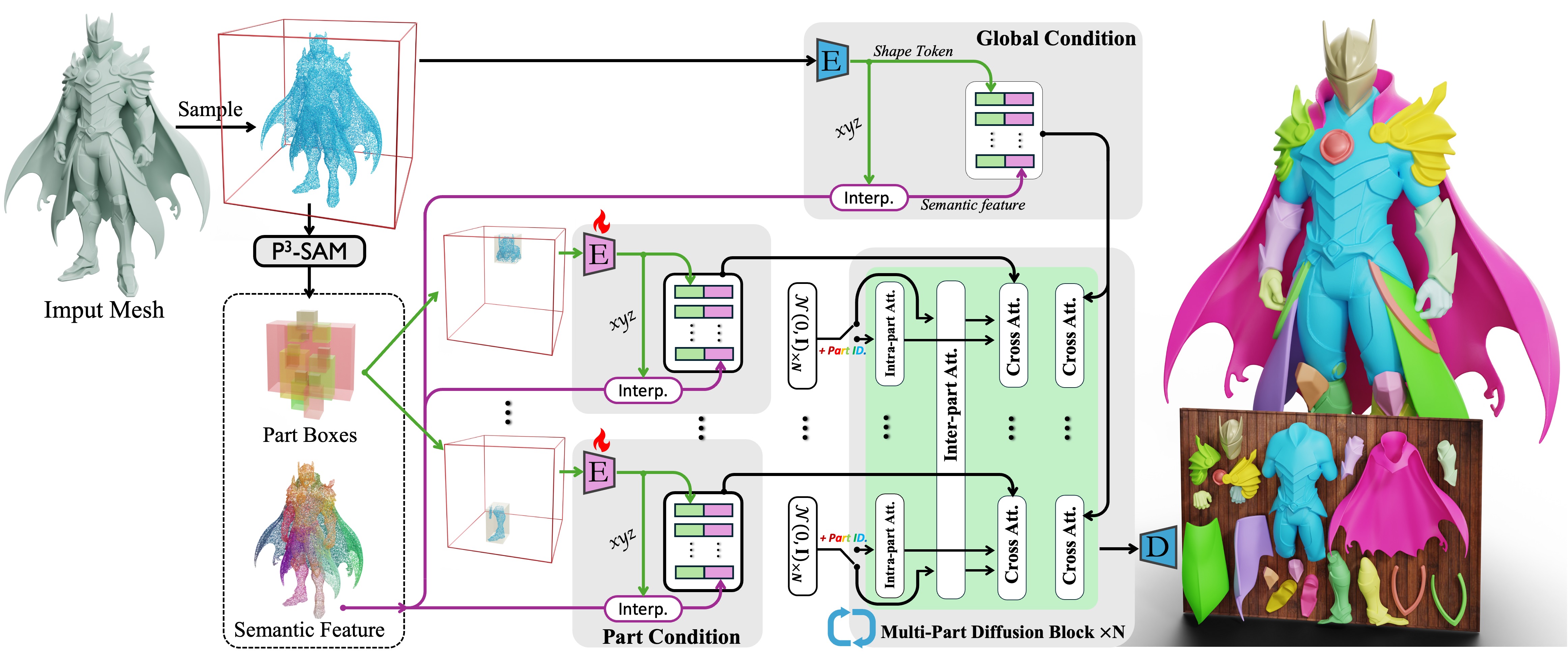}
  \caption{Pipeline of our shape decomposition.}
  \label{fig:xpart-pipeline}
\end{figure*}

Motivated by these observations, we introduce $\mathcal{X}$-Part, a controllable and editable diffusion framework, which enables semantically meaningful and structurally coherent part generation.
Our objective is to generate high-fidelity and structure-coherent part geometries from a given object point cloud, while ensuring flexible controllability over the decomposition process. 
Figure.~\ref{fig:xpart-pipeline} shows the pipeline of our shape decomposition method.

First, to achieve the controllability, we propose a part-level cues extraction module that uses bounding boxes as prompts to indicate part locations and scales, instead of directly using segmentation results as input. Compared with fine-grained and point-level segmentation cues, bounding boxes provide a coarser form of guidance, which mitigates overfitting to the input. Besides, the bounding box provides additional volume scale information for the partially visible part, benefiting the generation and controllability.

Second, despite inaccuracies in the segmentation results, we notice that the high-dimension point-wise semantic feature is free from the information compression caused by the cluster algorithm or prediction head used in~\cite{partfield2025}, resulting in more accurate semantic representations. Therefore, we carefully introduce the semantic features into our framework with delicately designed feature perturbation, which benefits the meaningful part decomposition. 

Third, we integrate $\mathcal{X}$-Part into a bounding box based part editing pipeline. It supports local editing, such as merging a small number of parts within an object and adjusting their scales, to facilitate interactive part generation. 
To prove the effectiveness of $\mathcal{X}$-Part, we conducted extensive experiments on various benchmarks. Our results show that $\mathcal{X}$-Part achieves state-of-the-art performance in part-level decomposition and generation. 

For more details of $\mathcal{X}$-Part, please refer to the paper~\cite{X-part}.

\begin{figure*}[!ht]
  \centering
  \includegraphics[width=1\linewidth]{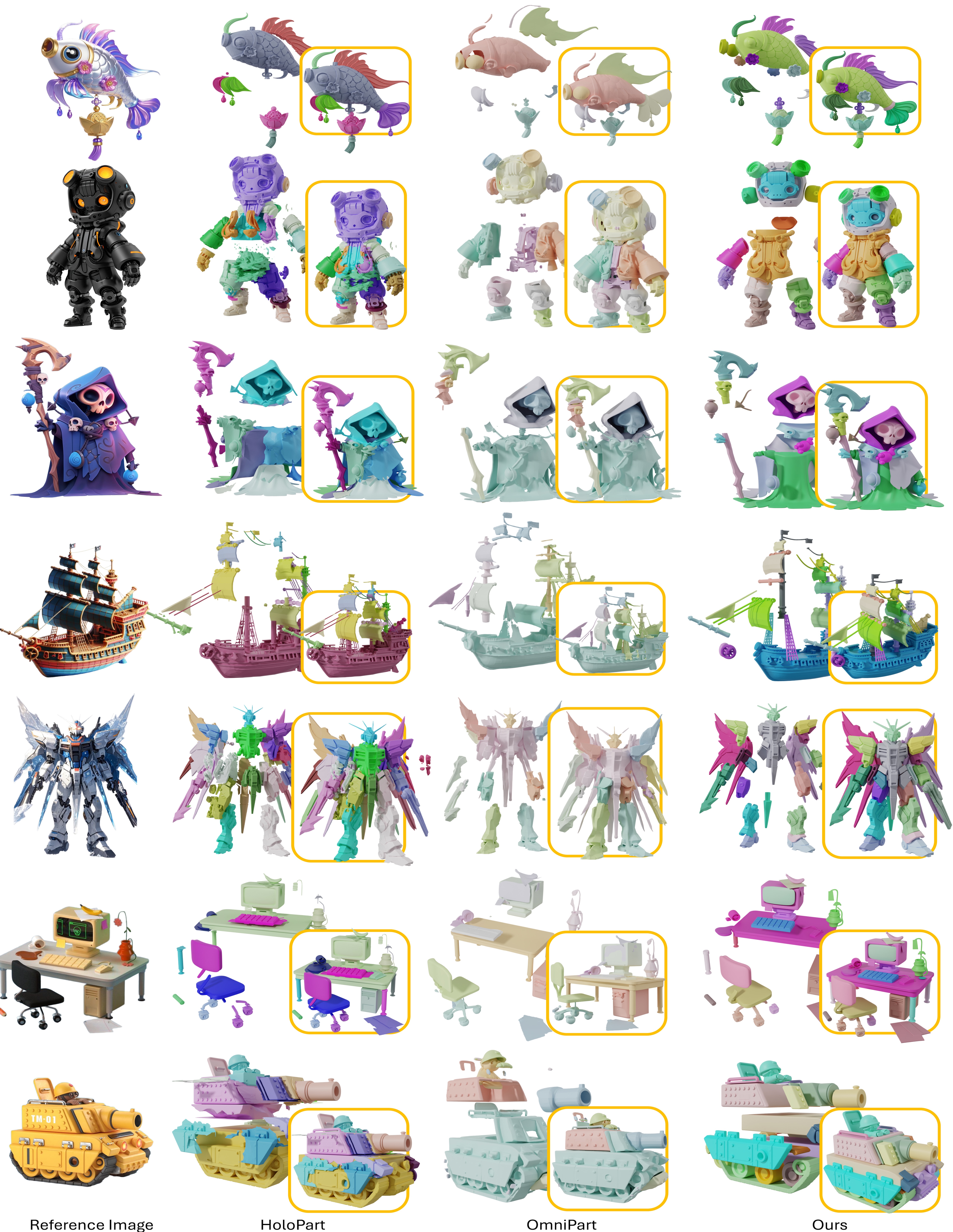}
  \caption{Shape Decomposition Results.}
  \label{fig:xpart-parts}
\end{figure*}

\textbf{Comparison with SOTA.}
We evaluate our method on 200 samples from the ObjaversePart-Tiny dataset, each comprising rendered images and corresponding ground-truth part geometries. To assess geometric quality, we employ Chamfer Distance (CD) and F-Score. The F-Score is computed at two different thresholds $[0.1, 0.5]$ to capture both coarse-level and fine-level geometric alignment. Prior to metric computation, each object is normalized to the range $[-1, 1]$. To ensure pose-agnostic evaluation, we rotate each object by $[0, 90, 180, 270]$ degrees and report the best score among these orientations as the final metric.
As shown in Table \ref{tab:xpart-tabel} and Figure \ref{fig:xpart-parts} our method outperforms all baselines.

\begin{table}[!b]
\centering
\label{tab:performance_comparison_3Dfront}
\begin{tabular}{lccccc}
\toprule
\textbf{Method} & CD↓ & Fscore-0.1↑ & Fscore-0.5↑ \\
\midrule
SAMPart3D  & 0. &  & 0. \\
PartField  & 0.17 & 0.68 & 0.57 \\
HoloPart  & 0.26 & 0.59 & 0.43  \\
OmniPart  & 0.23 & 0.63 & 0.46 \\
Ours      & \textbf{0.11} & \textbf{0.80} & \textbf{0.71}  \\
\bottomrule
\end{tabular}
\caption{Shape Decomposition Results}
\label{tab:xpart-tabel}
\end{table}

%% file: sections/polygen.tex
\section{Polygon Generation with Auto-regressive Models}

\begin{figure*}[th]
\centering
\vspace{-0.3mm}
\includegraphics[width=\linewidth]{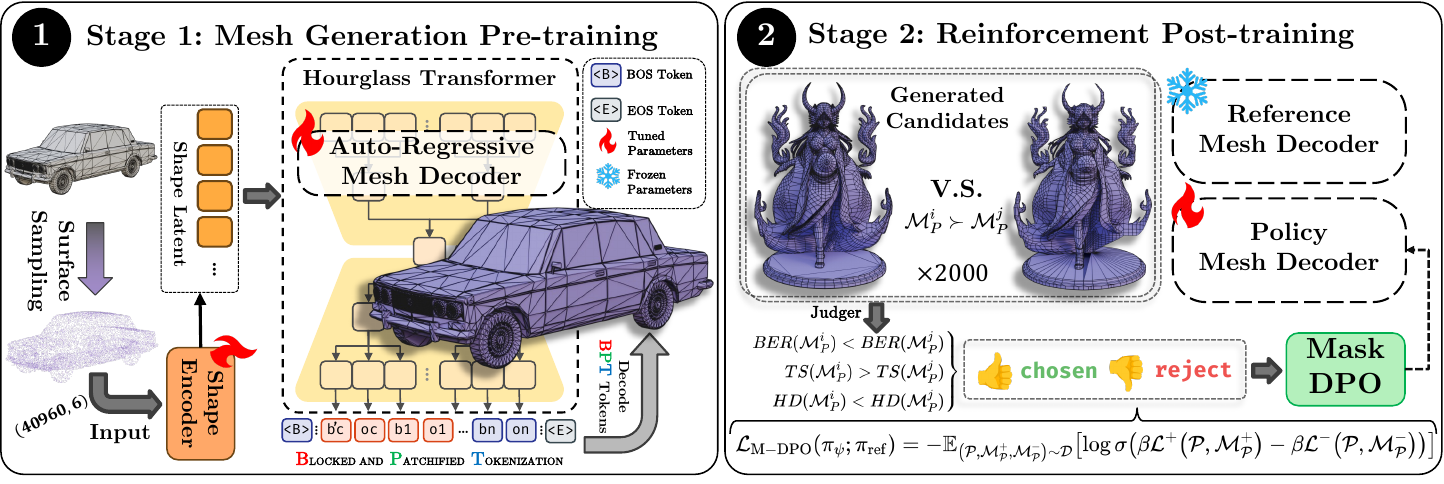}
\caption{\textbf{Mesh-RFT Framework Overview.} The pipeline comprises two stages: \textbf{1) Mesh Generation Pre-training} using an Hourglass AutoRegressive Transformer and a Shape Encoder;  and \textbf{2) Reinforcement Post-training} which employs Mask DPO with reference and policy networks for subsequent refinement.}
\label{fig:poly-pipeline}
\vspace{-5pt}
\end{figure*}

In this section, our goal is to generate a clean topology for shapes produced by geometry generative models or given by users. Despite the delicate shapes produced in previous sections, they typically consist of a huge amount of messy triangles and are hard to be directly applied in downstream applications (e.g., UV segmentation and rigging).
Therefore, we leverage an auto-regressive model to directly predict vertices and faces from the point cloud of generated shapes.
Our model is trained in two stages, including pre-training and post-training as shown in Figure \ref{fig:poly-pipeline}.

\subsection{Mesh Pretrianing with Efficient Tokenization and Architecture}

\paragraph{Mesh Tokenization.}

To model meshes with the next-token-prediction paradigm, the first step is to tokenize them into 1D sequences.
We leverage the Blocked and Patchified Tokenization (BPT) \citep{weng2025scaling} as the basic tokenization for meshes.
Specifically, BPT combines two core mechanisms: 1) \textit{Block-wise Indexing} partitions 3D coordinates into discrete spatial blocks, converting Cartesian coordinates into block-offset indexes to exploit spatial locality. 2) \textit{Patch Aggregation} further compresses face-level data by selecting high-degree vertices as patch centers and aggregating connected faces into unified patches. Each patch is encoded as a center vertex followed by peripheral vertices, reducing vertex repetition and enhancing spatial coherence.
With BPT, both training and inference efficiency are significantly improved.

\paragraph{Network Architecture.}
Our polygon generative model consists of a point cloud encoder and an auto-regressive mesh decoder.
The point cloud encoder is highly motivated by Michelangelo \citep{zhao2023michelangelo} and Hunyuan3D series \citep{zhao2025hunyuan3d,hunyuan3d2025hunyuan3d}, which applies the Perceiver \citep{jaegle2021perceiver} architecture to encode point cloud into condition tokens $c_p$. 
Then, we leverage the Hourglass Transformer \citep{hao2024meshtron} as the mesh decoder backbone, conditioned on point cloud tokens by cross attention layers.

\paragraph{Training \& Inference Scheme.}

The mesh token distribution $p(m_i)$ is modeled with Hourglass Transformer with parameter $\theta$, maximizing the log probability. The \textit{cross-attention} is leveraged for various conditions $c_p$.
\begin{equation}
    L(\theta) =  \prod_{i=1}^{|m|} p(m_{i} | m_{1:i-1}, c_p;\theta),
\end{equation}
To further leverage the high-poly mesh data and improve training efficiency, we adopt the truncated training strategy \citep{hao2024meshtron}.
Specifically, for each training iteration, we randomly selected a slice of mesh sequence with a fixed number of faces (e.g., 4k faces).
And in the inference stages, we apply the rolling cache strategy to reduce the gap between the training and inference stages.

\subsection{Mesh Post-Training with Topology-Aware Masked DPO}

\begin{figure*}[htbp]
    \centering
    \includegraphics[width=\linewidth]{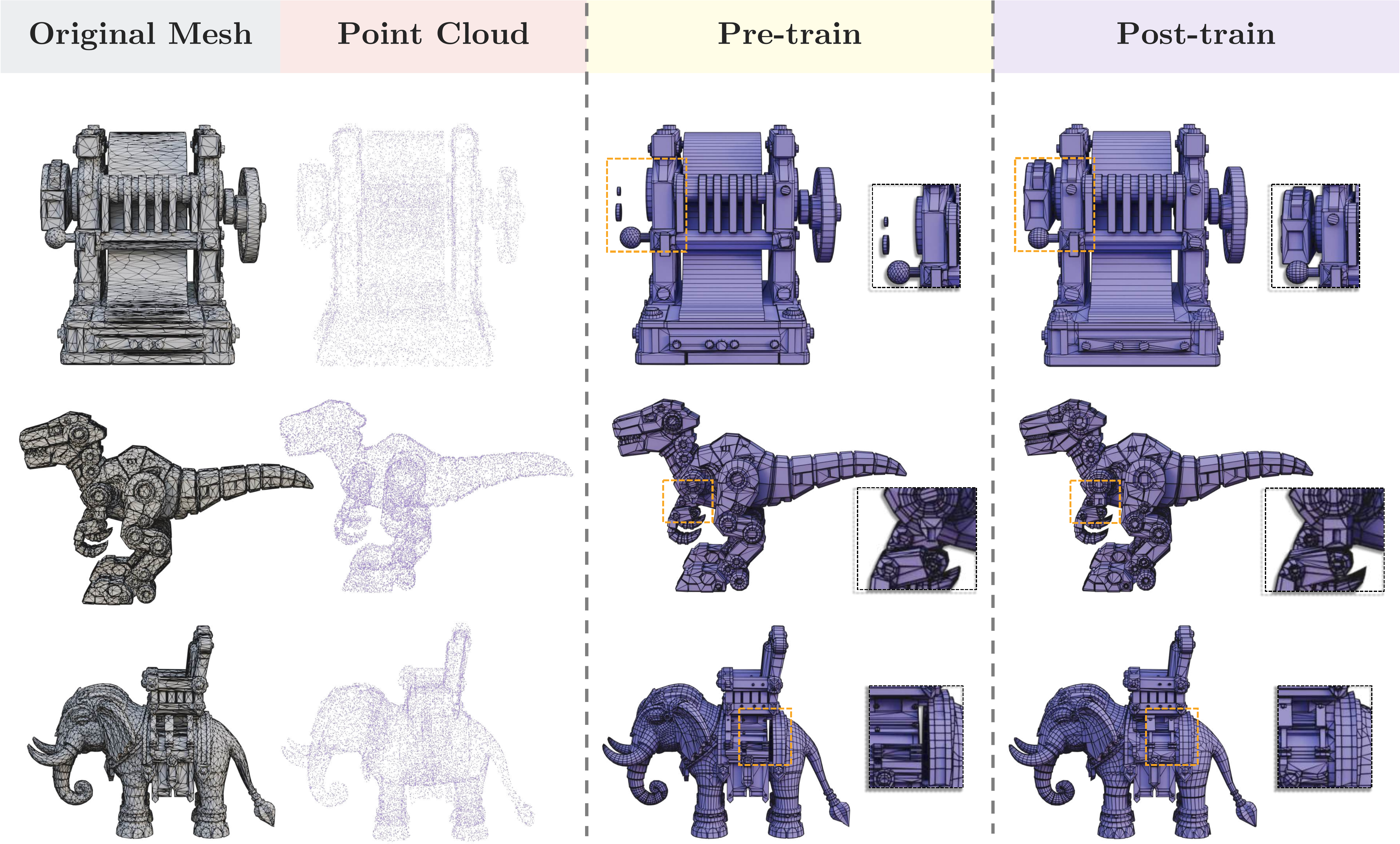}
    \caption{\textbf{The effectiveness of the post-training stage.} The post-training stage enhances the mesh completeness (Row \#1) and connectivity (Row \#2) and reduces the broken faces (Row \#3).}
    \label{fig:poly-prepost}
    \vspace{-0.1cm}
\end{figure*}

\paragraph{Preference Dataset Construction.}
We establish a pipeline for constructing the preference dataset used in the second-stage fine-tuning, which consists of candidate generation, multi-metric evaluation, and preference ranking.

For each input point cloud $\mathcal{P}$, we generate eight candidate meshes $\{\mathcal{M}_{\mathcal{P}}^1, \mathcal{M}_{\mathcal{P}}^2, \ldots, \mathcal{M}_{\mathcal{P}}^8\}$ using the pre-trained model $G_{\theta}^{\text{pre}}$. Each candidate is evaluated using three metrics: Boundary Edge Ratio (BER) and Topology Score (TS) for topological quality, and Hausdorff Distance (HD) for geometric consistency.

A preference relation $\mathcal{M}_{\mathcal{P}}^i \succ \mathcal{M}_{\mathcal{P}}^j$ is defined if and only if:
\begin{equation}
\begin{aligned}
\text{BER}(\mathcal{M}_{\mathcal{P}}^i) &< \text{BER}(\mathcal{M}_{\mathcal{P}}^j) \quad \land \\
\text{TS}(\mathcal{M}_{\mathcal{P}}^i) &> \text{TS}(\mathcal{M}_{\mathcal{P}}^j) \quad \land \\
\text{HD}(\mathcal{M}_{\mathcal{P}}^i) &< \text{HD}(\mathcal{M}_{\mathcal{P}}^j)
\end{aligned}
\end{equation}
We compile preference triplets $(\mathcal{P}, \mathcal{M}_{\mathcal{P}}^+, \mathcal{M}_{\mathcal{P}}^-)$ from all pairwise comparisons to form the dataset.

\paragraph{Masked Direct Preference Optimization.}
To address localized geometric imperfections and inconsistent face density, we leverage Masked Direct Preference Optimization (M-DPO) \citep{liu2025mesh}, which extends DPO with quality-aware localization masks.

We define a binary masking function $\phi(\mathcal{M}) \in \{0,1\}^{|\mathcal{M}|}$ that identifies high-quality regions (value 1) versus low-quality regions (value 0) based on per-face quality assessment. Each region corresponds to a subsequence in the block patch tokenization (BPT). A subsequence is classified as a high-quality region only if all faces within it have a quad ratio above a predefined threshold and the average topology score exceeds another threshold.

Let $G_{\text{ref}} = G_\theta^{\text{pre}}$ be the frozen reference model and $G_\psi$ be the trainable policy. The M-DPO objective is:
\begin{equation}
\mathcal{L}_{\text{M-DPO}}(\pi_\psi; \pi_{\text{ref}}) = -\mathbb{E}_{(\mathcal{P}, \mathcal{M}_{\mathcal{P}}^+, \mathcal{M}_{\mathcal{P}}^-) \sim \mathcal{D}} \left[ \log \sigma \left( \beta \mathcal{L}^+(\mathcal{P}, \mathcal{M}_{\mathcal{P}}^+) -\beta \mathcal{L}^-(\mathcal{P}, \mathcal{M}_{\mathcal{P}}^-) \right)\right]
\end{equation}
where the positive and negative terms are:
\begin{equation}
\begin{aligned}
\mathcal{L}^+(\mathcal{P}, \mathcal{M}_{\mathcal{P}}^+) &= \log \frac{\left|\pi_\psi(\mathcal{M}_{\mathcal{P}}^+ | \mathcal{P}) \odot \phi(\mathcal{M}_{\mathcal{P}}^+)\right|_1}{\left|\pi_{\text{ref}}(\mathcal{M}_{\mathcal{P}}^+ | \mathcal{P}) \odot \phi(\mathcal{M}_{\mathcal{P}}^+)\right|_1} \\
\mathcal{L}^-(\mathcal{P}, \mathcal{M}_{\mathcal{P}}^-) &= \log \frac{\left|\pi_\psi(\mathcal{M}_{\mathcal{P}}^- | \mathcal{P}) \odot (1-\phi(\mathcal{M}_{\mathcal{P}}^-))\right|_1}{\left|\pi_{\text{ref}}(\mathcal{M}_{\mathcal{P}}^- | \mathcal{P}) \odot (1-\phi(\mathcal{M}_{\mathcal{P}}^-))\right|_1}
\end{aligned}
\end{equation}
Here, $\odot$ denotes element-wise multiplication and $|\cdot|_1$ is the $\ell_1$ norm. M-DPO enables targeted refinement of low-quality regions while preserving satisfactory areas.

\subsection{Experiments}

\begin{figure*}[htbp]
    \centering
    \includegraphics[width=\linewidth]{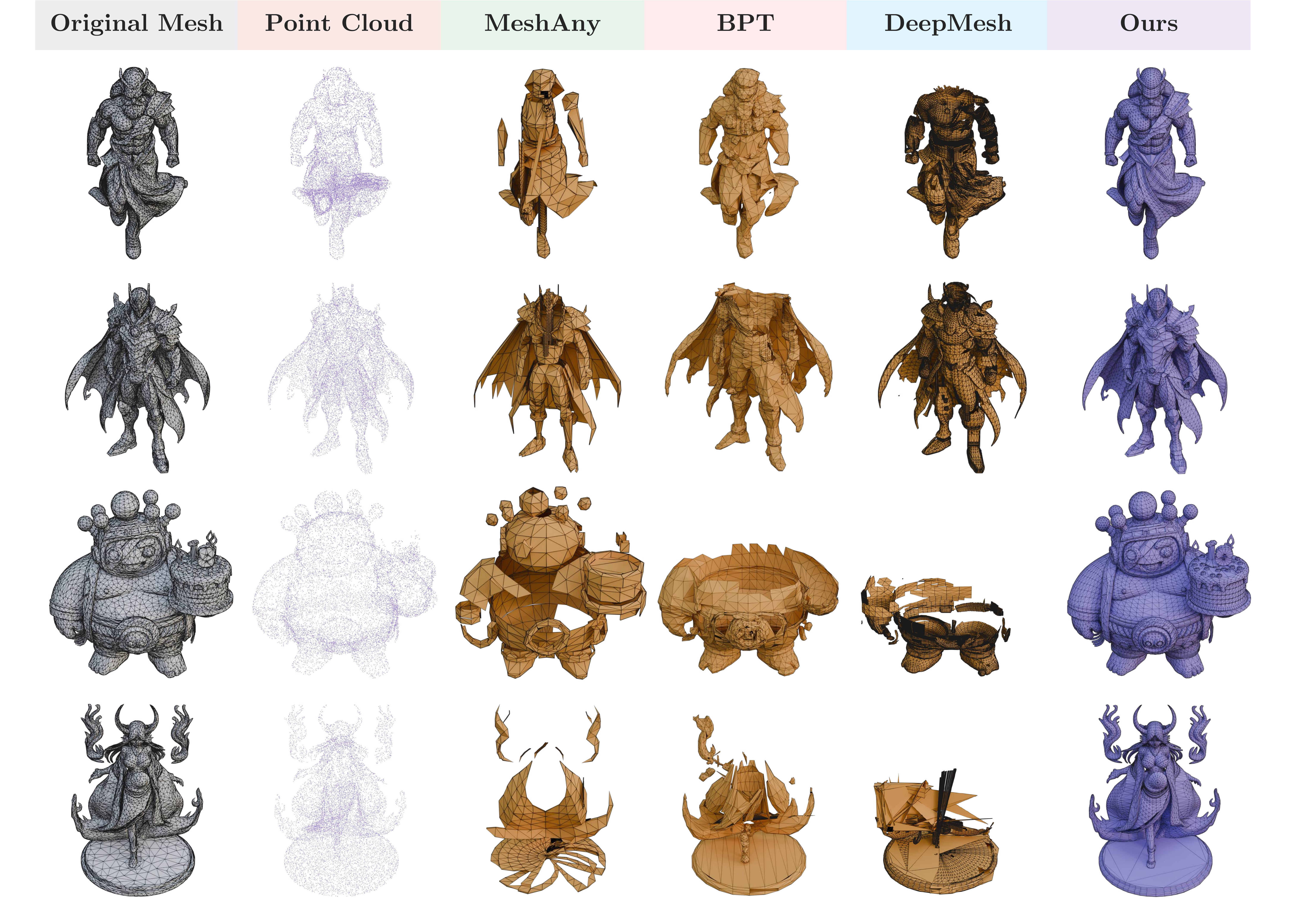}
    \caption{\textbf{Generalization results on dense, out-of-distribution meshes.} Our model demonstrates superior geometric fidelity and surface continuity, maintaining high-quality reconstruction even under complex and unseen input conditions.}
    \label{fig:poly-comparison}
    
\end{figure*}

\begin{figure*}[htbp]
    \centering
    \includegraphics[width=0.65\linewidth]{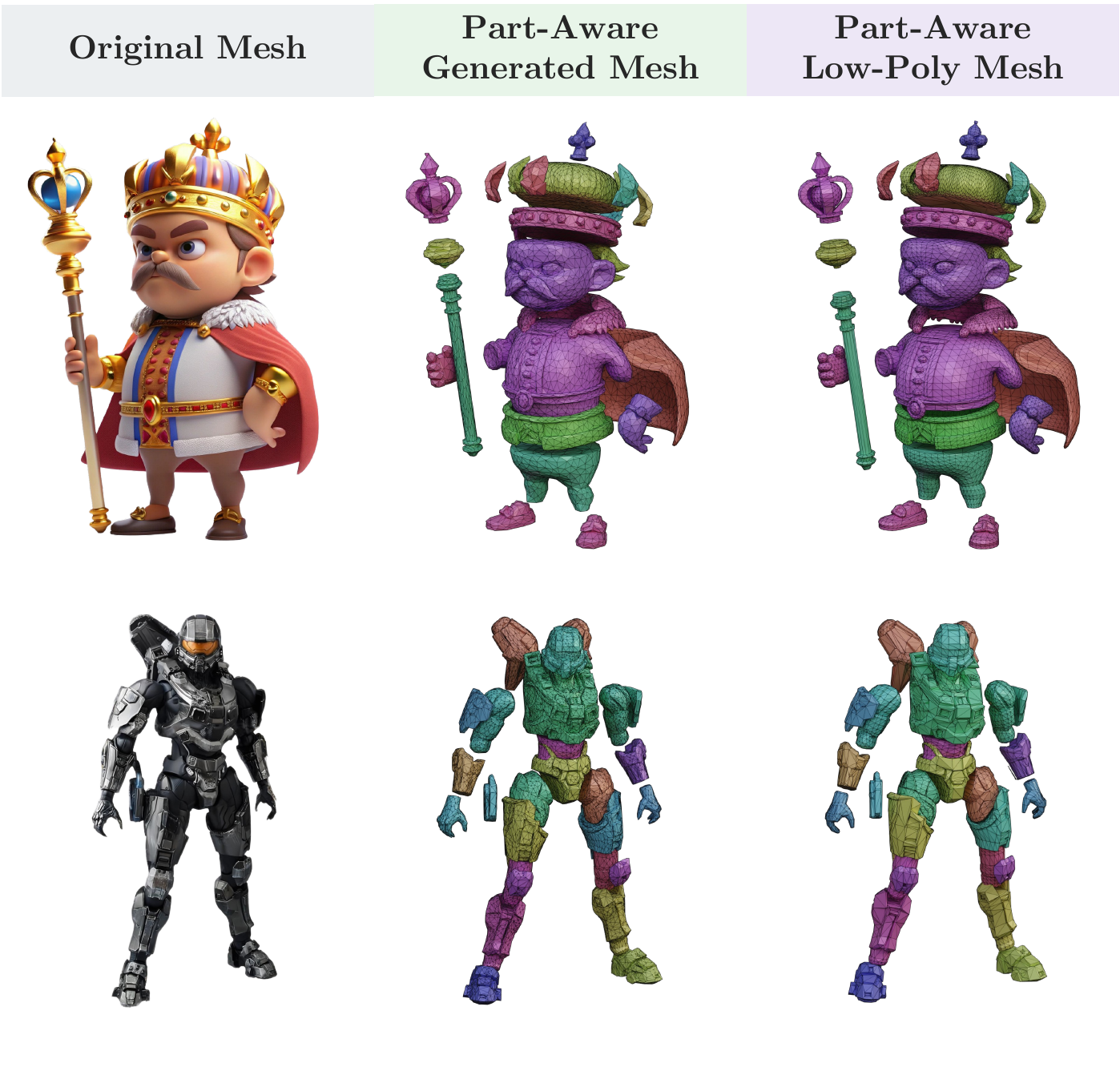}
    \caption{\textbf{Part-aware polygon generation.} With shapes segmented into several parts as input, our model can generate the corresponding meshes conditioned on partial point clouds separately without further fine-tuning. }
    \label{fig:poly-part}
\end{figure*}

\paragraph{Pre-training vs. post-training.} 
In Figure \ref{fig:poly-prepost}, we show the improvement after the post-training.
In our experiments, we found that the post-training stage is crucial for improving the completeness and topology quality of the generated meshes.

\paragraph{Comparison with existing methods.}

As shown in Figure \ref{fig:poly-comparison}, we compare our model with existing polygon generation methods.
Our model can generate much more complex meshes with significantly improved topology quality and stability.

\paragraph{Part-aware polygon generation.}

With shapes segmented into several parts as input, our model can generate the corresponding meshes conditioned on partial point clouds separately without further fine-tuning as shown in Figure \ref{fig:poly-part}.
This would be much easier for the model to generate the topology for complicated meshes.